\documentclass[journal]{IEEEtran}
\pdfoutput=1

\usepackage{amsmath,amssymb,amsthm}
\usepackage{graphicx}
\usepackage{booktabs}
\usepackage{multirow}
\usepackage{microtype}
\usepackage{xcolor}
\usepackage[numbers,sort&compress]{natbib}
\usepackage[colorlinks=true,linkcolor=blue!60!black,citecolor=blue!60!black,urlcolor=blue!60!black]{hyperref}

\newtheorem{theorem}{Theorem}
\newtheorem{corollary}{Corollary}
\newtheorem{proposition}{Proposition}
\newtheorem{lemma}{Lemma}
\theoremstyle{definition}
\newtheorem{definition}{Definition}

\theoremstyle{remark}

\newcommand{\tol}{\mathrm{tol}}
\newcommand{\ws}{w^{*}}
\newcommand{\dsing}{d}

\newcommand{\Rc}{\widehat{\mathcal{R}}_{\mathrm{c}}}
\newcommand{\Rt}{\widehat{\mathcal{R}}_{\mathrm{t}}}

\title{Chebyshev Polynomial Zonotope Abstractions for\\
Certified Reachability of Learning-Enabled Systems}

\author{Peng~Xie and Amr~Alanwar%
\thanks{Peng Xie and Amr Alanwar are with the TUM School of Computation, Information and Technology, Department of Computer Engineering, Technical University of Munich, 74076 Heilbronn, Germany (e-mail: p.xie@tum.de; alanwar@tum.de).}%
\thanks{This work is funded by the Deutsche Forschungsgemeinschaft (DFG, German Research Foundation), project number 564740977.}%
\thanks{Code, all raw results, pre-registered prediction notes, and reproduction scripts (cluster job identifiers included) accompany the paper.}}

\begin{document}
\maketitle

\begin{abstract}
Nearly all current methods for computing reachable sets of
nonlinear dynamics propagate them with truncated Taylor expansions
plus rigorous remainders. The expansion is pointwise: once its
argument is a whole set rather than a point, one polynomial must
stay accurate across that set, and it reaches only as far as its
disc of convergence. The limit binds hardest on the rational terms
and sigmoid gates of modern controllers: smooth as these look on
the real line, in the complex plane they blow up at points just off
the real axis, closer as a gate sharpens, and those points cap the
set width a Taylor expansion can cover, most tightly on the
switching surface where the controller decides. We prove a
dichotomy: no Taylor degree certifies sets wider than that cap,
while a Chebyshev abstraction of equal degree, anchored to the
whole set in the same polynomial-zonotope pipeline, has no cap: its
certified width grows linearly with the degree while its error falls
at a geometric rate. Measured Taylor walls converge to the predicted cap from above,
within 2 percent by degree 32, and the Chebyshev wall crosses it at
degree 20. We give closed-form rigorous
remainders, among them, to our knowledge, the first polynomial
softmax transformer; existing sound treatments are linear
relaxations. On closed loops,
certificates survive where the Taylor ones diverge, and
specifications flip from \textsc{unknown} to \textsc{proven}; on
trained classifiers up to real MNIST digits, the Chebyshev frontier
certifies input territory the Taylor basis surrenders. Every
certified claim passed large Monte-Carlo harnesses with zero
violations, and an applicability map states in advance where the
basis provably cannot matter.
\end{abstract}


\section{Introduction}\label{sec:intro}

Set-based reachability analysis asks whether \emph{all} trajectories
from a \emph{set} of initial states satisfy a specification, which no
finite batch of simulations can establish. The mature tools answer
the question in the same way. Flow* \citep{chen2013flowstar}, CAPD
\citep{kapela2021capd} and JuliaReach
\citep{bogomolov2019juliareach} propagate Taylor-model
flowpipes; CORA polynomializes conservatively over (sparse)
polynomial zonotopes
\citep{althoff2013reachability,kochdumper2021sparse}; POLAR pushes
polynomial arithmetic through neural-network loops
\citep{huang2022polar}. In every case, the nonlinear
functions on the
current enclosure are replaced by a truncated \emph{Taylor} expansion
plus a rigorous remainder, and every tool that entered the ARCH-COMP
nonlinear category in the last five years is Taylor-based
\citep{geretti2021arch}.

The default is easy to defend. A truncated Taylor series is the
best possible \emph{local} model: exact to order $N$ at its
expansion point, and unbeatable over a narrow enclosure. The
mismatch appears when the enclosure is wide, because set propagation
asks one polynomial to be accurate uniformly across the whole set, a
different approximation problem from accuracy at a point, and one
that a pointwise expansion answers only inside its disc of
convergence. A Chebyshev expansion is anchored to the whole
set instead of to its center, which is exactly the request set
propagation makes. This paper is about the class of functions
where that mismatch bites hardest, and about how little it costs to
fix it. The class
covers the rational expressions of vehicle and orbital dynamics
($1/(1+x^2)$-type damping, $\mu/r^3$ gravity) and, more importantly
for modern control, the \emph{gates}, meaning the sigmoid and softmax
nonlinearities that blend controller modes in gain scheduling
\citep{rugh2000research}, mixture-of-experts policies
\citep{shazeer2017outrageously}, and smooth approximations of switching
logic. What all of these have in common is complex poles close to the
real axis; for a gate, the pole distance shrinks in proportion to its
sharpness.

\begin{figure*}[t]
\centering
\includegraphics[width=\textwidth]{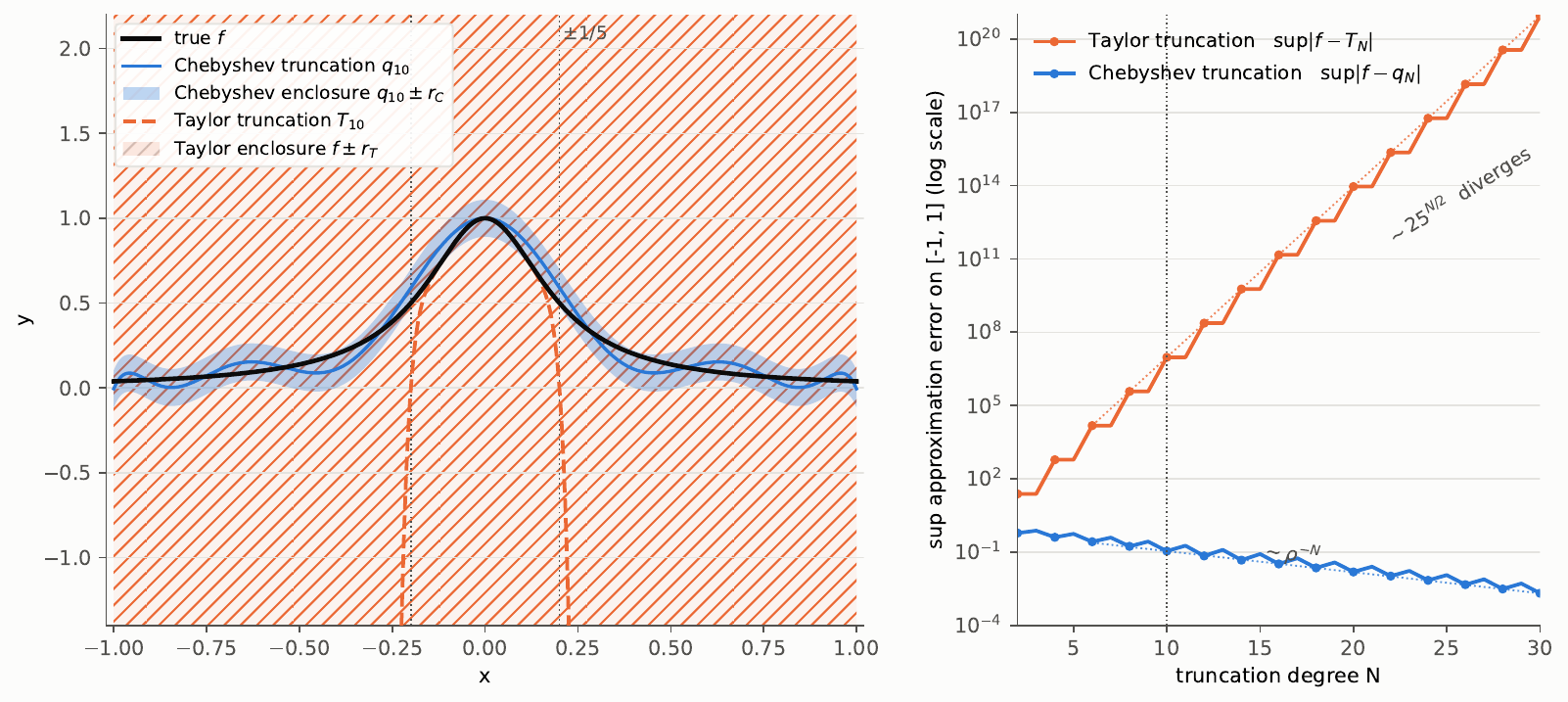}
\caption{The Runge function $g(x)=1/(1+25x^2)$ on $[-1,1]$ at
degree $N=10$: shipped enclosures (left) and sup error versus
degree (right).}
\label{fig:runge1d}
\end{figure*}

Figure~\ref{fig:runge1d} shows the mismatch in its simplest form.
Both enclosures have exactly the structure a polynomial-zonotope
pipeline carries, a degree-$10$ polynomial plus a remainder
interval; they differ only in \emph{which} polynomial is shipped.
The Taylor enclosure comes out $3\times10^{7}$ times wider than the
truth, and raising the degree makes it exponentially worse; the
Chebyshev enclosure is $1.3$ times the truth and improves
geometrically. The mechanism, developed in
Section~\ref{sec:theory}, is classical approximation theory inside
a certification pipeline. Taylor convergence lives on a \emph{disc},
which has to cover the whole enclosure while avoiding the poles.
Chebyshev convergence needs a \emph{Bernstein ellipse}, and the
ellipse only has to keep the poles off the real interval itself. For
a gate of temperature $\tau$, the poles sit at height $\pi\tau$,
so sharper gates pull them in, and practice is moving toward sharper
gates.

\paragraph{Contributions.}
\begin{enumerate}
\item \textbf{A predictive criterion} (Section~\ref{sec:theory}): a
dichotomy theorem (Taylor's certified half-width is capped at
$\dsing\cdot\tol^{1/N}$; Chebyshev's is uncapped, at a degree growing
linearly with the width)
with two switching-surface corollaries: the \emph{temperature law}
($\dsing=\sqrt{m^2+\pi^2\tau^2}$ for sigmoid gates) and the worst-case corollary (the
certified radius attains its minimum exactly above the switching
surface).
Measured Taylor walls converge to the cap from above (within $2\%$
by $N=32$) and the Chebyshev wall crosses it at $N\approx20$; a scope
proposition delimits where the basis provably does \emph{not} matter.
\item \textbf{A drop-in abstraction} (Section~\ref{sec:method}):
the set representation stays sparse polynomial zonotopes exactly as
shipped, and only the basis of the scalar abstraction changes:
Chebyshev coefficients with closed-form sound remainder families
(several new, including singularity-aware bounds), under a
family-minimum discipline: every call evaluates several
independently sound bounds and ships the tightest, symmetrically for
both bases, including the classical interval-Lagrange member, which
on Chebyshev nodes gains a further free factor $2^{N}$ from the
monic minimality of the nodal polynomial; to our knowledge no
reachability method ships its remainders this way, and measured
scalar-for-scalar it tightens the standard interval-Lagrange recipe
at every tested width (Figure~\ref{fig:family}). Plus a sound softmax transformer built on the
log-sum-exp route, the first polynomial one (linear-relaxation
treatments exist \citep{shi2020transformers,bonaert2021deept}).
\item \textbf{Certified consequences} (Section~\ref{sec:experiments}):
propagation certificates that survive where Taylor's diverge; safety
specifications that flip from \textsc{unknown} to \textsc{proven};
certified softmax decision radii that Chebyshev reaches at degree $1$ and
Taylor only at degree $8$; full-plane
certification maps on trained classifiers, up to a spiral-boundary
network where the basis change alone recovers five to seven points of
input-plane area and an MNIST classifier certified against
brightness and contrast changes where interval arithmetic certifies
nothing; a quantified
``just split the set'' tax; and an external anchor that shows our
Taylor baseline is no strawman.
\item \textbf{A predictive applicability map}
(Section~\ref{sec:limits}): the criterion forecasts its own null cases
(polynomial dynamics tie bit-for-bit; entire nonlinearities gain
constant factors), and Monte-Carlo attribution locates the high-degree
ceiling in the box representation rather than in the dynamics.
\end{enumerate}

All experiments were pre-registered in analysis notes before the runs.
Every certified set is checked, step by step, against dense
Monte-Carlo and corner batteries; the total violation count across
the paper is zero.


\section{Related work}\label{sec:related}

This section reviews the prior work closest to each ingredient of
the paper.

\paragraph{Taylor-based set propagation.} Taylor models with interval
remainders originate with Berz and Makino \citep{berz1998verified} and
power the flowpipe tools Flow* \citep{chen2013flowstar} and CAPD
\citep{kapela2021capd}, with validated-integration roots in
\citet{lohner1987,nedialkov1999validated}. CORA's conservative
polynomialization \citep{althoff2013reachability} and sparse polynomial
zonotopes \citep{kochdumper2021sparse} carry Taylor coefficients in a
set representation; POLAR \citep{huang2022polar} extends the approach to
neural-network controllers, and Verisig \citep{ivanov2019verisig} maps
sigmoid networks to hybrid automata, with Taylor-model
preconditioning in Verisig 2.0 \citep{ivanov2021verisig2}. The
set-representation lineage behind our pipeline is the zonotope one:
\citet{girard2005zonotope} for linear systems, order-reduction
analyses in \citet{kopetzki2017methods}, and the
constrained-zonotope and ellipsotope generalizations
\citep{scott2016constrained,kousik2022ellipsotopes}; interval
foundations trace to \citet{moore2009interval}, Taylor-model
integrator performance is examined in \citet{neher2007taylor}, and the
wider hybrid-systems tooling includes SpaceEx, dReach and CORA
\citep{frehse2011spaceex,kong2015dreach,althoff2015cora}.
Validated Runge--Kutta integrators bound the step defect directly
\citep{sandretto2016validated}; affine arithmetic ships minimax
(Chebyshev-type) degree-one enclosures of elementary functions
\citep{defigueiredo2004affine}, a linear precedent for the basis
choice made here; and \citet{bompadre2013convergence} analyze the
convergence orders of Taylor and McCormick--Taylor models.

On the closed-loop neural-network side, ReachNN certifies with
Bernstein polynomials \citep{huang2019reachnn}, Sherlock with
regressive polynomial rules \citep{dutta2019sherlock}, NNV with star
sets \citep{tran2019star,tran2020nnv}, OVERT with sound
piecewise-linear overapproximations \citep{sidrane2022overt}, and
\citet{schilling2022verification} integrate Taylor models with
zonotopes;
the ARCH-COMP AINNCS category collects these tools annually
\citep{johnson2021arch}. All are Taylor-based where they meet smooth
dynamics; ARCH-COMP nonlinear reports contain no non-Taylor
expansion basis \citep{geretti2021arch}.

\paragraph{Open-loop neural-network verification.} The
robustness-certification line runs from SMT-complete solvers
\citep{katz2017reluplex,katz2019marabou} through
abstract-interpretation domains \citep{gehr2018ai2,singh2019deeppoly}
and linear bound propagation \citep{zhang2018crown,wang2021betacrown},
with verified training \citep{gowal2018ibp,wong2018provable},
branch-and-bound on top \citep{bunel2020branch}, a survey in
\citet{liu2021algorithms}, and an annual competition
\citep{brix2023vnncomp}. Transformer certification bounds attention
with linear relaxations \citep{shi2020transformers,bonaert2021deept},
and semantic perturbation families, brightness, contrast, geometry,
are certified by relaxation in
\citet{mohapatra2020semantify,balunovic2019deepg}. These methods
target piecewise-linear or per-neuron-relaxed networks; the window
where a shared-symbol polynomial method pays in that world, and
where it provably cannot, is measured in
Section~\ref{sec:nnwindow}.

\paragraph{Chebyshev approximation in validated computing.} Chebyshev
series with rigorous remainders appear in computer-assisted proofs
\citep{brisebarre2010chebyshev,vandenberg2015rigorous}, in validated
integration of single trajectories \citep{dzetkulic2015rigorous}, in
validated Chebyshev spectral methods for linear ODEs
\citep{brehard2018validated}, in Chebyshev model arithmetic for
global optimization \citep{rajyaguru2017chebyshev}, which does
propagate set-valued enclosures of factorable functions, and,
historically, in ultra-arithmetic \citep{epstein1982ultra}.
\citet{arioli2026comparative} recently compared validated Taylor and
Chebyshev integrators on single solutions. None of these lines
integrates with a zonotopal shared-symbol representation, and none
addresses controllers or closed loops, so
to our knowledge the combination Chebyshev + polynomial zonotopes +
reachability is new. The same calibration applies to softmax: sound
linear-relaxation transformers exist
\citep{shi2020transformers,bonaert2021deept}; what is new here is a
polynomial, shared-symbol treatment with rigorous sup-norm
remainders that does not collapse on wide logit sets. We do not
consider rational (Pad\'e) abstractions: polynomial zonotopes are
not closed under rational maps, so a quotient would leave the
representation, and with it the exact polynomial arithmetic the
pipeline is built on.


\section{Problem Setup and Preliminaries}
\label{sec:setup}

This section fixes the setting that both bases share.

\subsection{Reachability with certificates}\label{sec:problem}

The object of verification in this paper is a discrete-time system
\begin{equation}\label{eq:dynamics}
x_{k+1} \;=\; \Phi(x_k),
\qquad k=0,1,\dots,K-1 ,
\end{equation}
whose transition map $\Phi:\mathbb{R}^n\to\mathbb{R}^n$ is a finite
composition of arithmetic operations and univariate primitives
(rational terms, sigmoid and $\tanh$ gates, and so on). Declaring
the discrete map as the model follows the standard practice of
discrete-time nonlinear reachability and MPC
\citep{bravo2006robust,althoff2015cora}, whose benchmark models are
themselves typically one-step discretizations of continuous plants.
Two sources supply such maps. In a sampled-data learning-enabled loop, $\Phi$
is one control period of the plant model advanced by a fixed
one-step integrator, composed with a neural controller evaluated
exactly, layer by layer: the map a simulator executes. And any
smooth system $\dot x=f(x)$ built from the same primitives yields
such a $\Phi$ through a one-step scheme of step size $h$;
throughout this paper the scheme is the classical fourth-order
Runge--Kutta method
\begin{equation}\label{eq:rk4}
\Phi_h(x)\;=\;x+\tfrac{h}{6}\,(\kappa_1+2\kappa_2+2\kappa_3+\kappa_4),
\end{equation}
with $\kappa_1=f(x)$, $\kappa_2=f(x+\tfrac{h}{2}\kappa_1)$,
$\kappa_3=f(x+\tfrac{h}{2}\kappa_2)$,
$\kappa_4=f(x+h\,\kappa_3)$. The point of writing this out is
structural: $\Phi_h$ is a finite composition of the arithmetic of
$f$ itself, so it is again of the declared form, and any one-step
scheme with this property would do. Discretization leaves a gap
between $\Phi_h$ and the time-$h$ flow map $\varphi_h$ of $f$,
where $\varphi_h(x)$ is the solution of $\dot x=f(x)$ at time $h$
started at $x$ ($f$ locally Lipschitz, solutions assumed to exist
through each step on the sets considered). Per step, a validated
enclosure
$\mathcal{E}_{\mathrm{flow}}^{(k)}\supseteq
\{\varphi_h(x)-\Phi_h(x):x\in\widehat{\mathcal{R}}_k\}$, with
$\widehat{\mathcal{R}}_k$ the enclosure constructed below, is
precisely what validated Runge--Kutta integrators construct
\citep{sandretto2016validated}. Added to the propagation
\eqref{eq:onestep} as one more Minkowski summand, it extends
every certificate from $\Phi$ to the continuous flow \emph{at
the sampling instants} $t=kh$; covering the times inside a step
takes one further ingredient, an intra-step time-interval
enclosure, which the same integrators provide. Both ingredients
are independent of the abstraction basis and identical for the
two methods, so omitting them biases neither side of the
comparison, and the flow defect is of order $h^{5}$ (the local
order of the scheme), far below the abstraction errors measured
here. The certificates of this paper are statements about $\Phi$,
with the continuous-time extension priced as above.

Starting from a set of initial states
$\mathcal{R}_0\subset\mathbb{R}^n$, given as a box or any
polynomial zonotope, the reachable sets of \eqref{eq:dynamics}
are defined exactly by
\begin{equation}\label{eq:reach}
\mathcal{R}_{k+1} \;=\; \Phi(\mathcal{R}_k) \;=\; \{\,\Phi(x)\;:\;x\in \mathcal{R}_k\,\}.
\end{equation}
For polynomial $\Phi$ the pipeline needs no abstraction at all:
polynomial zonotopes are closed under polynomial maps, every step
stays in the class, and the two methods coincide bit for bit
(Section~\ref{sec:ablation}). The moment $\Phi$ contains a
transcendental primitive this closure is lost, and provably so: a
polynomial zonotope is the image of a box under a polynomial map,
hence a semialgebraic set by the Tarski--Seidenberg theorem, while
already the planar set $\{(x,\sigma(x)):x\in[0,1]\}$, the graph
of a sigmoid over an interval, is not semialgebraic (else
$e^{-x}=(1-\sigma)/\sigma$ would be algebraic over
$\mathbb{R}(x)$). No object of the class equals such an image, at
any degree, so $\mathcal{R}_k$ can only be enclosed from outside.
The pipeline therefore replaces each transcendental primitive by a
polynomial together with a \emph{rigorous error bound} on the
whole argument range; polynomial operations introduce no further
approximation (closure of the class), and the error bound is added
back as a set, so the approximation and its certified error term
together form a guaranteed overapproximation: writing $q_k$ for
the polynomial replacement of $\Phi$ anchored on the current
enclosure and $\mathcal{E}_k$ for its certified error set (both
depend on the step and on the basis),
\begin{equation}\label{eq:onestep}
\widehat{\mathcal{R}}_{k+1} \;:=\;
q_k\big(\widehat{\mathcal{R}}_{k}\big)\;\oplus\;\mathcal{E}_{k}
\;\supseteq\; \Phi\big(\widehat{\mathcal{R}}_{k}\big),
\qquad \widehat{\mathcal{R}}_{0}=\mathcal{R}_{0},
\end{equation}
with $\oplus$ the Minkowski sum. Soundness follows by induction:
$\widehat{\mathcal{R}}_0=\mathcal{R}_0$, and if
$\widehat{\mathcal{R}}_k\supseteq\mathcal{R}_k$ then
$\widehat{\mathcal{R}}_{k+1}\supseteq
\Phi(\widehat{\mathcal{R}}_k)\supseteq\Phi(\mathcal{R}_k)
=\mathcal{R}_{k+1}$, the middle step because set images are
monotone under inclusion. The \emph{enclosure}
$\widehat{\mathcal{R}}_k$ therefore satisfies the soundness
invariant
\begin{equation}\label{eq:soundness}
\widehat{\mathcal{R}}_k \;\supseteq\; \mathcal{R}_k
\qquad\text{for all } k=0,\dots,K .
\end{equation}
When the abstraction basis matters we tag the enclosure with it and
write $\Rc^{(k)}$ for the Chebyshev method and $\Rt^{(k)}$ for the
Taylor method; both are instances of $\widehat{\mathcal{R}}_k$. The hat always
marks a certified over-approximation of the true set $\mathcal{R}_k$, and $k$
indexes control steps up to the horizon $K$. The truncation degree of the
scalar abstraction is written $N$ throughout, and method names carry
it: Taylor-$4$ means $N=4$.

A specification asks the enclosures to remain in an invariant set
and/or to reach a goal set. The pipeline returns \textsc{proven}
when the enclosures establish this, and \textsc{unknown} otherwise;
by \eqref{eq:soundness} the answer \textsc{proven} is sound, while
\textsc{unknown} carries no information, occurring both for unsafe
systems and for safe systems whose enclosures are too conservative.

The quality question of this paper is therefore
\emph{conservatism}, measured by enclosure volume: the tighter the
enclosures, the more specifications the same method can prove, and
excess conservatism turns \textsc{proven} into \textsc{unknown}.

Every certificate in this paper is a statement about the map
$\Phi$ itself. Within the semantics
\eqref{eq:dynamics}--\eqref{eq:reach} nothing is missing from
\eqref{eq:onestep}: the abstraction error $\mathcal{E}_k$ is the
only error there is, and it is displayed and added at every step;
the continuous-time extension is priced with \eqref{eq:rk4}. The
one comparison against a flow-level tool is read with this
asymmetry stated (Section~\ref{sec:anchor}).

\subsection{Polynomial zonotopes}\label{sec:pz}

Every enclosure in this paper is carried in one representation.

\begin{definition}[Polynomial zonotope \citep{kochdumper2021sparse}]
\label{def:pz}
Given a center $c\in\mathbb{R}^{n}$, dependent generators
$G\in\mathbb{R}^{n\times h}$ with exponent matrix
$E\in\mathbb{N}_{0}^{p\times h}$, and independent generators
$G_{I}\in\mathbb{R}^{n\times q}$, the associated \emph{polynomial
zonotope} is the set
\begin{equation}\label{eq:pz}
\mathcal{PZ}
=\Big\{\,
\begin{aligned}
c&+\sum_{i=1}^{h}\Big(\prod_{l=1}^{p}\alpha_{l}^{E_{(l,i)}}\Big)
G_{(\cdot,i)}
+\sum_{j=1}^{q}\beta_{j}\,G_{I(\cdot,j)}\\[-2pt]
&\;:\;\alpha_{l},\beta_{j}\in[-1,1]
\Big\}.
\end{aligned}
\end{equation}
\end{definition}

The dependent part carries dependencies between state coordinates
through the shared factors $\alpha$; the independent part is an
ordinary zonotope. Our pipeline instantiates this with dependent
exponents graded up to a total degree $D$, an independent pool of $P$
columns, and rigorous scalar remainders stored as an interval box,
itself a set of axis-aligned independent generators.

\subsection{The scalar abstraction}
\label{sec:abstraction}

Propagating \eqref{eq:reach} through this representation requires
three operations: (i)~\emph{exact} PZ arithmetic for sums and
products of sets of the form \eqref{eq:pz}
\citep{kochdumper2021sparse}; (ii)~order reduction back to degree
$D$ and pool size $P$
\citep{girard2005zonotope,kopetzki2017methods}; and (iii)~the only
nonlinear step, a \emph{univariate scalar abstraction}. Given a primitive
$g:\mathbb{R}\to\mathbb{R}$ (e.g.\ $\sigma$, $\tanh$, $1/(1+x^{2})$)
and the interval enclosure $[m-w,\,m+w]$ of its PZ argument, the
abstraction of degree $N$ and basis $B\in\{\text{Taylor},
\text{Chebyshev}\}$ returns a polynomial--radius pair
\begin{equation}\label{eq:abstraction}
\begin{aligned}
&\mathcal{A}^{B}_{N}\;:\;\big(g,\,[m-w,\,m+w]\big)\;\longmapsto\;(q,\,r),\\
&\quad\sup_{|x-m|\le w}\big|g(x)-q(x)\big|\;\le\;r ,
\end{aligned}
\end{equation}
after which $g$ is replaced by $q$ composed with the argument PZ, and
$r$ is added to the remainder radius. The pair $(q,r)$ is the entire
interface between the basis and the shared machinery. Operations
(i) and (ii) are standard; every result of this paper concerns the
choice of basis in (iii), and Section~\ref{sec:method} constructs
both instantiations.

\subsection{The ablation principle}\label{sec:ablation}

Soundness makes each method trustworthy on its own; it does not
yet make the two comparable. The comparison in this paper is a controlled
experiment: the two
methods share the representation, the reduction step, the step size
and the semantics, and differ in a single operation. This rests on one
structural fact about the machinery of Section~\ref{sec:abstraction}.

\begin{lemma}[Ablation principle]\label{lem:ablation}
Consider two methods that share the representation \eqref{eq:pz}, the
arithmetic and reduction steps (i)--(ii), the step size, the horizon,
and the tolerance, and differ only in the basis $B$ of the scalar
abstraction \eqref{eq:abstraction}. Then:
\begin{enumerate}
\item[(a)] \emph{Soundness is basis-independent.} The invariant
\eqref{eq:soundness} holds for any abstraction satisfying the
inequality in \eqref{eq:abstraction}; the basis influences only the
size of the returned radius $r$.
\item[(b)] \emph{Polynomial dynamics are a structural tie.} If $f$ is
polynomial, step (iii) is never invoked, and the two methods return
identical enclosures $\widehat{\mathcal{R}}_k$, bit for bit, in floating-point
arithmetic.
\item[(c)] \emph{Any separation is attributable.} Consequently, every
difference observed between the two methods on non-polynomial $f$ is
caused by the approximation quality of the shipped pair $(q,r)$, and by
nothing else.
\end{enumerate}
\end{lemma}

\begin{proof}
(a) Enclosure propagation composes set-wise sound operations; the
inequality in \eqref{eq:abstraction} is the only property of $(q,r)$
used in the soundness argument. (b) For polynomial $f$, every
right-hand side is built from sums and products, which steps
(i)--(ii) evaluate identically in both methods; the basis $B$
enters only through \eqref{eq:abstraction}, which a polynomial
right-hand side never invokes. (c) is immediate from (a) and
(b).
\end{proof}

Claim (b) is verified numerically in Section~\ref{sec:limits}: over
280 paired runs on Van der Pol and Duffing systems the maximal
deviation between the two methods is exactly $0.0$.


\section{The certified-wall dichotomy: Taylor divergence beyond the
singularity distance, geometric Chebyshev convergence}
\label{sec:theory}

This section proves the main theory: a dichotomy between the
certified widths of the two methods, two switching-surface laws,
and the provable tie cases. We first define the quantities the
results bound. Fix a tolerance $\tol$. The \emph{scalar certified
wall} of basis $B$ is the largest argument half-width whose
remainder the abstraction certifies to tolerance,
\begin{equation}\label{eq:scalar-wall}
\ws_{B}(N;m)\;=\;\max\big\{\,w \;:\; r\le\tol \text{ in }
\eqref{eq:abstraction}\,\big\}.
\end{equation}
The \emph{system certified wall} $\ws$ is the largest initial-set
half-width the full method carries to the horizon. Operationally,
a run \emph{certifies} half-width $w$ if every step's enclosure
radius stays below a fixed divergence threshold (failure is
detected as radius overflow or a non-finite value before step $K$)
and the specification check passes at the horizon; $\ws$ is the
supremum of certified $w$. Certification was monotone in $w$ on
every grid we measured, so both walls are located by bisection.

\subsection{Disc versus ellipse}\label{sec:geometry}

Let $g$ be analytic on a neighborhood of $X=[m-w,\,m+w]$ with nearest
complex singularity at distance $\dsing$ from $m$. The degree-$N$ Taylor
truncation $T_N$ at $m$ converges only on the open disc of radius
$\dsing$: for every $\rho$ with $w<\rho<\dsing$,
\begin{equation}\label{eq:taylor-rate}
\sup_{X}|g-T_N| \;\le\; C_{\rho}\,(w/\rho)^{N}
\end{equation}
with $C_{\rho}=C_{\rho}(g,w)$ independent of $N$, and $\rho=\dsing$
itself is admissible when the nearest singularity is a pole, the case
of every primitive used here. For $w>\dsing$ the error grows
geometrically in $N$ (Theorem~\ref{thm:dichotomy}(i) makes the
constants explicit): the disc must cover the whole enclosure, and it
must avoid the singularity. The Chebyshev
interpolant $q_N$ on $X$ instead obeys the Bernstein-ellipse bound
\citep{trefethen2019atap}
\begin{equation}\label{eq:cheb-rate}
\sup_{X}|g-q_N| \;\le\; \frac{4M(\rho)}{\rho-1}\,\rho^{-N}
\qquad\text{for every } 1<\rho<\rho^{*},
\end{equation}
where $E_\rho$ is the ellipse with foci $m\pm w$ and parameter $\rho$,
$M(\rho)=\sup_{E_\rho}|g|$, and $\rho^{*}$ is the parameter at which the
ellipse first touches a singularity. The ellipse degenerates to the
interval as $\rho\to1$: it needs the singularity \emph{off the real
interval}, not outside a disc. For a pole at height $\dsing$ above the
interval, $\rho^{*}\ge \dsing/w+\sqrt{1+\dsing^2/w^2}>1$ for every
$w$: Chebyshev never diverges; it only pays degree.

\begin{theorem}[Certified-wall dichotomy]\label{thm:dichotomy}
Fix a tolerance $\tol$. Let the nearest singularities of $g$ be a
complex-conjugate pair of simple poles at distance $\dsing$ from
$m$, off the real axis, with every other singularity at distance
$\dsing_{2}>\dsing$, and let $w<\dsing_{2}$. Then:
\emph{(i)} there are constants $c_{g}>0$ and $N_{0}$, depending
only on $g$ and $w/\dsing$, such that for every $N\ge N_{0}$ a
sound Taylor abstraction of degree $N$ admits a certificate
$r\le\tol$ only if
\[
w \;\le\; \dsing\cdot\big(N\tol/c_{g}\big)^{1/(N+1)}
\;=\;\dsing\cdot\tol^{1/N}\,\big(1+o(1)\big),
\]
so $\ws_{T}(N)\le\dsing\cdot\tol^{1/N}(1+o(1))$: no degree
certifies any width beyond the singularity distance;
\emph{(ii)} the Chebyshev abstraction admits a certificate for
\emph{every} $w$ once
\[
N \;\ge\; \frac{\log\!\big(4M(\rho)\,/\,((\rho-1)\,\tol)\big)}
{\log\rho}
\qquad\text{for any } \rho<\rho^{*}(w),
\]
a degree logarithmic in $1/\tol$ and, for $w\gg\dsing$, linear in
$w/\dsing$ up to a logarithmic factor from $M(\rho)$.
\end{theorem}

The pole-pair hypothesis covers every gate and rational primitive
in this paper (the pole locations are computed before
Corollaries~\ref{cor:temperature}--\ref{cor:valley}); poles of
higher order only strengthen part (i). Part (ii) uses none of this
structure: analyticity in a Bernstein ellipse suffices, for any
singularity type.

The cap itself needs no pole structure either. If $g$ has radius of
convergence $\dsing$ at $m$, whatever the singularity type,
Cauchy--Hadamard gives $\limsup_k|a_k|^{1/k}=1/\dsing$ for the Taylor
coefficients, so for $w>\dsing$ the terms $a_k w^{k}$ are unbounded
and $\sup_X|g-T_N|$ is unbounded in $N$: for every $w>\dsing$ the
certified wall falls below $w$ for infinitely many degrees. The pole
hypothesis is what turns this unboundedness into the explicit rate
$(w/\dsing)^{N+1}/N$ of part (i).

\begin{proof}
\emph{(i) The Taylor bound.} Write the nearest singularities of $g$ as
a complex-conjugate pair of simple poles $\zeta,\bar\zeta$ with
$\zeta-m=\dsing e^{i\theta}$, $\theta\in(0,\pi)$; this covers every
instance in this paper ($\sigma$, $\tanh$, and the rational terms
have simple poles), and poles of higher order only strengthen the
bound below by polynomial factors in $N$. Extract the principal
part,
\[
g(x)=2\,\mathrm{Re}\!\left[\frac{A}{x-\zeta}\right]+h(x),
\]
where $h$ is analytic up to the second-nearest singularity distance
$\dsing_2>\dsing$ (for the gates, $\dsing_2$ is the distance to the
poles at three times the height). For the pole pair the degree-$N$
Taylor tail is an exact geometric remainder: with
$u=(x-m)/(\zeta-m)$,
\[
g(x)-T_N(x)=2\,\mathrm{Re}\!\left[\frac{A}{m-\zeta}\,
\frac{u^{\,N+1}}{1-u}\right]+\big(h-T_N[h]\big)(x),
\]
and $\sup_X|h-T_N[h]|\le C_2\,(w/\dsing_2)^{N+1}$ by the Cauchy
bound, a strictly smaller geometric base. Now bound the pole-pair
term from below on the sub-segment
$x=m+s\dsing$, $s\in[(1-1/N)\,w/\dsing,\;w/\dsing]$. There
$u=s\,e^{-i\theta}$: the modulus of the bracket is at least
$\frac{|A|}{\dsing(1+w/\dsing)}\,s^{N+1}
\ge \frac{|A|}{\dsing(1+w/\dsing)}\,e^{-2}\,(w/\dsing)^{N+1}$
(since $(1-1/N)^{N+1}\ge e^{-2}$ for $N\ge3$; smaller $N$ is
absorbed into $N_0$ below), while its argument is
$\varphi_0-(N+1)\theta-\arg(1-u)$, and
$s\mapsto\arg(1-s e^{-i\theta})$ is strictly monotone with derivative
bounded below by $\sin\theta/(1+w/\dsing)^{2}$. Over the sub-segment
the argument therefore sweeps an interval of width
$\Delta_N\ge c_1(\theta,w/\dsing)/N>0$, and an interval of width
$\Delta$ always contains a point with
$|\cos(\cdot)|\ge\sin(\Delta/2)\ge\Delta/\pi$. Combining,
\begin{align*}
\sup_X|g-T_N| &\ge \frac{c_g}{N}\,(w/\dsing)^{N+1}
-C_2\,(w/\dsing_2)^{N+1}\\
&\ge \frac{c_g'}{N}\,(w/\dsing)^{N+1}
\end{align*}
for all $N\ge N_0(g,w)$. A sound degree-$N$ abstraction must return
$r\ge\sup_X|g-T_N|$, so $r\le\tol$ forces
$(c_g'/N)(w/\dsing)^{N+1}\le\tol$, that is
\begin{align*}
w&\le\dsing\cdot\tol^{1/(N+1)}\,(N/c_g')^{1/(N+1)}\\
&=\dsing\cdot\tol^{1/N}\,\big(1+o(1)\big),
\end{align*}
since $\tol^{1/(N+1)}=\tol^{1/N}\exp(\log(1/\tol)/(N(N+1)))$ and
$(N/c_g')^{1/(N+1)}\to1$; the constant $c_{g}$ of the statement is
$c_g'$. For
$w>\dsing$ the same lower bound grows without bound in $N$: beyond
the singularity distance no tolerance is certifiable at large degree,
and raising the degree makes the enclosure exponentially worse.

\emph{(ii) The Chebyshev degree bound.} Map the interval to the unit circle
by the Joukowski substitution $x=m+\tfrac{w}{2}(v+v^{-1})$; the
Bernstein ellipse $E_\rho$ is the image of $|v|=\rho$. A singularity
at height $\dsing$ vertically above $m$ lies on $E_{\rho^{*}}$ with
semi-minor axis $\tfrac{w}{2}(\rho^{*}-1/\rho^{*})=\dsing$, giving
$\rho^{*}=\dsing/w+\sqrt{1+\dsing^{2}/w^{2}}>1$ for every $w$; any
singularity off the closed interval gives $\rho^{*}>1$ likewise.
Inequality \eqref{eq:cheb-rate} is the Chebyshev interpolation bound
of \citet[Thm.~8.2]{trefethen2019atap}. Solving
$4M(\rho)\rho^{-N}/(\rho-1)\le\tol$ for $N$ gives the stated degree
bound, logarithmic in $1/\tol$. For $w\gg\dsing$,
$\rho^{*}=1+\dsing/w+O\big((\dsing/w)^{2}\big)$, so
$\log\rho^{*}=(\dsing/w)(1+o(1))$ and the required degree grows
linearly in $w/\dsing$; no width becomes uncertifiable.
\end{proof}

In summary: the certified width of the Taylor method never exceeds
$\dsing$, at any degree, and past $\dsing$ a larger degree makes
the certified radius exponentially \emph{worse}; we call this the
\emph{degree inversion} (Fig.~\ref{fig:runge1d}, right). The
Chebyshev method certifies every width, at a degree growing
linearly in $w/\dsing$. At the degree used throughout the
experiments, $N=4$, the operative difference is not the asymptotic
cap but the rate at $w\approx\dsing$: Taylor's $(w/\dsing)^{N}$ is $1$
there while the Chebyshev rate $(1+\sqrt2)^{-N}$ is $0.03$, so the
remainder injected where a closed loop decides is smaller by more
than an order of magnitude (the true errors of the sigmoid and
$1/(1+x^{2})$ primitives differ by $22\times$ and $41\times$ at that
width).

\subsection{Switching-surface laws}\label{sec:laws}

For gates, $\dsing$ is a design parameter. The logistic gate
$\sigma(x/\tau)$, ubiquitous in smooth gain scheduling and neural
policies, has poles at $x=\pm i\pi\tau$ (nearest to the real axis);
$\tanh(x/\tau)$ at $\pm i\pi\tau/2$.

\begin{corollary}[Temperature law]\label{cor:temperature}
For $g=\sigma(x/\tau)$ abstracted on $[m-w,m+w]$,
\[
\ws_{T}(N)\;=\;\sqrt{m^{2}+\pi^{2}\tau^{2}}\;\tol^{1/N}\,(1+o(1)),
\qquad N\to\infty,
\]
The certified radius of the Taylor abstraction collapses linearly with
the gate sharpness $1/\tau$; for $\tanh$ the constant $\pi$ is replaced
by $\pi/2$.
\end{corollary}

\begin{corollary}[Worst case at the switching surface]\label{cor:valley}
As a function of the distance $m$ of the enclosure center to the
switching surface, $\ws_{T}(N;m)$ has its global minimum exactly at
$m=0$: the Taylor abstraction is \emph{most} restricted directly above
the switch, the point a verification task cares about most, and
recovers only as $\sqrt{m^{2}+\pi^{2}\tau^{2}}$.
\end{corollary}

\begin{proof}[Proof of Corollaries~\ref{cor:temperature}
and~\ref{cor:valley}]
The gate $\sigma(z/\tau)=1/(1+e^{-z/\tau})$ is singular exactly where
$e^{-z/\tau}=-1$, that is at $z=i\pi\tau(2n{+}1)$, $n\in\mathbb{Z}$.
These are simple poles: the derivative of the denominator,
$-e^{-z/\tau}/\tau=1/\tau$, does not vanish there. The pair nearest
to a real center $m$ is $\pm i\pi\tau$, at distance
$\sqrt{m^{2}+\pi^{2}\tau^{2}}$; every other pole is at distance at
least $\sqrt{m^{2}+9\pi^{2}\tau^{2}}$, strictly farther, so the
hypotheses of Theorem~\ref{thm:dichotomy}(i) hold with
$\dsing=\sqrt{m^{2}+\pi^{2}\tau^{2}}$ and the cap follows by
substitution. For $\tanh(z/\tau)$ the singularities solve
$e^{2z/\tau}=-1$, giving simple poles at $z=i\pi\tau(2n{+}1)/2$ and
the constant $\pi/2$ in place of $\pi$. Corollary~\ref{cor:valley} is the shape
of the substituted cap: $m\mapsto\sqrt{m^{2}+\pi^{2}\tau^{2}}$ is
strictly increasing in $|m|$ with minimum $\pi\tau$ at $m=0$, so at
every fixed degree the cap is smallest directly above the switch and
grows like $|m|$ once $|m|\gg\pi\tau$.
\end{proof}

\begin{figure*}[t]
\centering
\includegraphics[width=\textwidth]{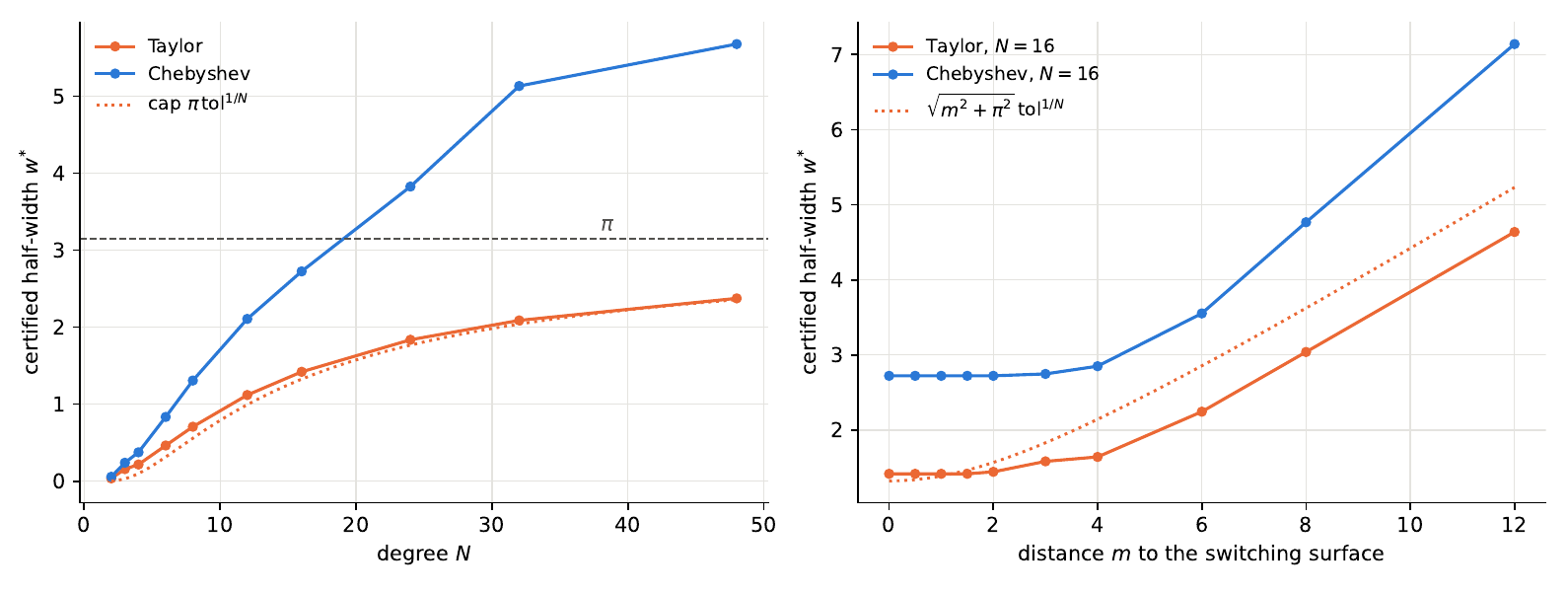}
\caption{Certified scalar walls (tolerance $10^{-6}$, sigmoid gate,
$\tau=1$), measured with the closed-form rigorous remainders of
Section~\ref{sec:method}. \emph{Left:} on the switching surface ($m=0$)
the measured Taylor wall converges to the bound $\pi\cdot\tol^{1/N}$
(Thm.~\ref{thm:dichotomy}) from above, at $2.2$, $1.26$, $1.07$,
$1.02$ and $1.007$ times it for $N=4,8,16,32,48$, and stays below
$\pi$; the Chebyshev wall passes $\pi$ at $N\approx19$ and keeps
growing ($1.63\pi$ at $N=32$, $1.81\pi$ at $N=48$). \emph{Right:} the
worst case of Cor.~\ref{cor:valley} at $N=16$: the Taylor wall attains
its minimum $1.42$ directly above the switch and follows
$\sqrt{m^2+\pi^2}\,\tol^{1/N}$ (dotted; $1.07\times$ it at $m=0$,
$0.89\times$ at $m=12$); the Chebyshev floor is $1.9\times$ higher.}
\label{fig:laws}
\end{figure*}

Figure~\ref{fig:laws} confronts both corollaries with measured certified
walls: the Taylor walls converge to the cap from above, at $2.2$,
$1.26$, $1.07$, $1.02$ and $1.007$ times $\pi\,\tol^{1/N}$ for
$N=4,8,16,32,48$ (the excess is the $(1+o(1))$ factor of
Theorem~\ref{thm:dichotomy}(i)), the $\tanh$ walls are $2^{-1-1/N}$
times the sigmoid walls to within $3\%$ ($0.489$ against $0.489$ at
$N=32$), which is the range factor of $\tanh(x)=2\sigma(2x)-1$ at an
absolute tolerance, and the per-degree divergence rate of the
Taylor abstraction in the rational family equals the predicted ratio
$w/\dsing$ to three decimals ($1/\rho^{*}=0.618$, the golden ratio, at
$w/\dsing=2$). This quantitative agreement is the paper's central evidence: the
failure is a law, and the law can be evaluated \emph{before} running a
tool.

\subsection{Example: the sigmoid at the switch}\label{sec:example}

The numbers behind Theorem~\ref{thm:dichotomy} and
Corollary~\ref{cor:temperature} can be written out on one gate.
Take $g=\sigma(x)$ on $[-4,4]$, the setting of
Table~\ref{tab:threebasis}: the switch is centered ($m=0$), the
poles sit at $\pm i\pi$, so $\dsing=\pi<w=4$ and the stress is
$w/\dsing=1.27$. At degree $N=10$ the two abstractions return
(odd terms; both expansions are odd around the switch apart from the
constant $\tfrac12$):
\begin{center}\small
\begin{tabular}{lrrr}
\toprule
$j$ & $b_j$ (Taylor) & $|b_j|\,4^{j}$ & $c_j$ (Chebyshev)\\
\midrule
$1$ & $2.5\times10^{-1}$ & $1.00$ & $5.59\times10^{-1}$\\
$3$ & $-2.08\times10^{-2}$ & $1.33$ & $-9.44\times10^{-2}$\\
$5$ & $2.08\times10^{-3}$ & $2.13$ & $2.15\times10^{-2}$\\
$7$ & $-2.11\times10^{-4}$ & $3.45$ & $-5.04\times10^{-3}$\\
$9$ & $2.14\times10^{-5}$ & $5.60$ & $1.13\times10^{-3}$\\
\bottomrule
\end{tabular}
\end{center}
The Taylor coefficients themselves decay, but the quantity that
enters the enclosure is the term size at the edge of the set,
$|b_j|w^{j}$, and that grows by the factor $(w/\dsing)^{2}=1.62$
per two degrees: the divergence of part (i) is visible inside the
coefficient table before anything is evaluated. The Chebyshev
coefficients contract instead, at the measured ratio
$|c_9/c_7|=0.224$ against the predicted
$(\rho^{*})^{-2}=0.236$ from
$\rho^{*}=\dsing/w+\sqrt{1+\dsing^{2}/w^{2}}=2.057$. The
resulting sup errors are $3.46$ for Taylor, $3.6\times$ the entire
range of $\sigma$ on the set, against $4.4\times10^{-4}$ for
Chebyshev, within the rate prediction
$(\rho^{*})^{-10}=7.4\times10^{-4}$. Sharpening the gate to
$\sigma(x/\tau)$ moves the poles to $\pm i\pi\tau$ and shrinks
$\dsing$ in proportion: Corollary~\ref{cor:temperature} is this
table repeated along the temperature axis.

\subsection{The certification norm is $L^\infty$: Chebyshev is
near-minimax, $L^2$-optimal and Bernstein bases are not}
\label{sec:basis}

The law rules Taylor out on this class; it does not by itself rule
Chebyshev in among the alternatives. What decides between the
candidates is the norm. Certification needs \emph{worst-case}
($L^\infty$) bounds: a remainder
radius is a sup-norm statement. Families that are optimal in $L^2$
(Legendre on an interval, Zernike polynomials on a disc
\citep{noll1976zernike}) minimize \emph{mean-square} error and provide
no worst-case guarantee at truncation; they target the wrong norm for a
certificate.
Bernstein polynomials, chosen in POLAR \citep{huang2022polar} for their
shape-preservation, converge only at rate $O(1/N)$ for Lipschitz
functions. Chebyshev interpolation is near-minimax in $L^\infty$: within
a factor $2+\tfrac{2}{\pi}\log N$ of the best possible polynomial
\citep{trefethen2019atap}, with the geometric rate
\eqref{eq:cheb-rate}. Table~\ref{tab:threebasis} makes the point on a
sharp gate in the divergence regime: at $N=64$, Taylor has diverged by
six orders of magnitude, Bernstein has bought one digit per degree
doubling, Chebyshev sits at machine precision.

\begin{table}[t]
\centering\footnotesize\setlength{\tabcolsep}{3.5pt}
\caption{Sup approximation error of three bases for $\sigma(x)$
on $[-4,4]$ (pole distance $\pi<4$).}
\label{tab:threebasis}
\begin{tabular}{rccc}
\toprule
$N$ & Taylor & Bernstein & Chebyshev\\
\midrule
4  & $8.2\times10^{-1}$ & $9.2\times10^{-2}$ & $3.1\times10^{-2}$\\
8  & $2.1\times10^{0}$  & $5.9\times10^{-2}$ & $1.8\times10^{-3}$\\
16 & $1.5\times10^{1}$  & $3.5\times10^{-2}$ & $5.9\times10^{-6}$\\
32 & $7.0\times10^{2}$  & $1.9\times10^{-2}$ & $5.8\times10^{-11}$\\
64 & $1.6\times10^{6}$  & $1.0\times10^{-2}$ & $5.2\times10^{-14}$\\
\bottomrule
\end{tabular}
\end{table}

\begin{figure}[t]
\centering
\includegraphics[width=\columnwidth]{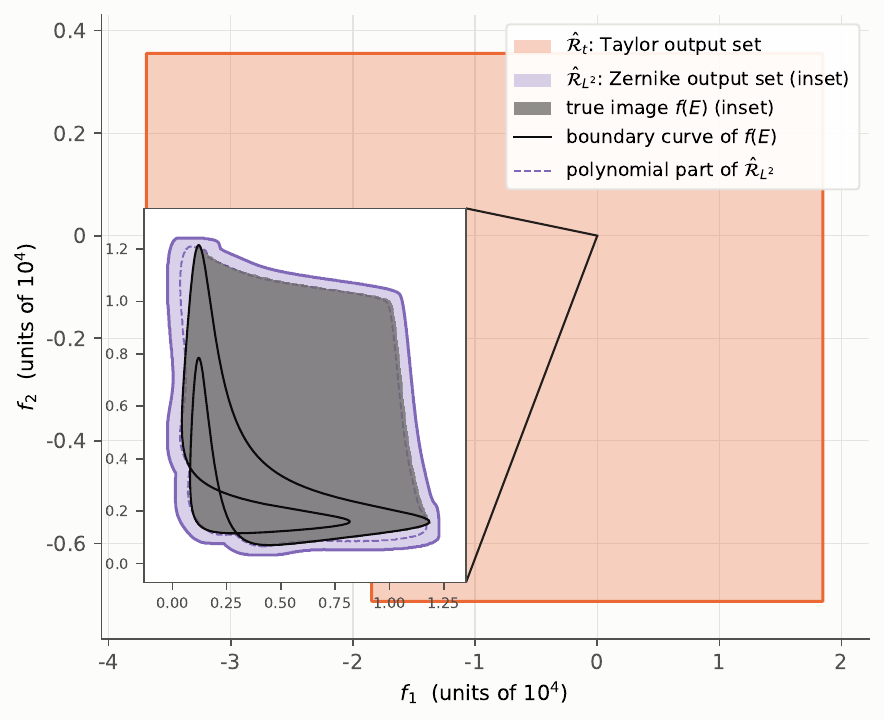}
\caption{The $L^{2}$ aside: an ellipsotope input (unit-disk parameter
domain, same center and generators as Figure~\ref{fig:runge2d})
through the same rational map, Taylor versus the total-degree-10
Zernike projection, the $L^{2}$-optimal polynomial on the disk. The
Zernike output set (inset) has area $1.22\times$ the true image with
measured sup remainder $r_{L^{2}}\approx(0.056,0.032)$; the Taylor
set spans $5\times10^{8}\times$ the truth.}
\label{fig:l2}
\end{figure}

Zernike and Chebyshev are built for different norms, and the aside
prices both in the only norm a certificate accepts. A method may
ship any polynomial, but the remainder it must attach is always that
polynomial's \emph{sup} error, so an $L^{2}$-optimal choice does not
escape the $L^{\infty}$ requirement. Figure~\ref{fig:l2} runs the test on
the family's home ground, an \emph{ellipsotope}
\citep{kousik2022ellipsotopes} (unit-disk parameter
domain) through the rational map of Figure~\ref{fig:runge2d}, with
the total-degree-10 Zernike projection, the ball analogue of
Legendre. The cost turns out modest, area $1.22\times$ the truth
next to Chebyshev's $1.17\times$ on the cube version of the same
map, because $L^{2}$ projections of analytic maps also converge
geometrically in sup norm; the sets do not explode. What the family
lacks is the certificate: its sup remainder must be measured after
the fact, with no analogue of the ellipse bound \eqref{eq:cheb-rate}
to guarantee it in advance and no closed-form sound remainder family
like those of Section~\ref{sec:method}. Chebyshev is the basis for
which the $L^{\infty}$ guarantee exists off the shelf.

\subsection{Provable ties: polynomial, entire, and real-axis regimes}
\label{sec:scope}

The criterion cuts both ways: the same geometry that predicts the
wins marks out where nothing can be won.

\begin{proposition}[Scope]\label{prop:scope}
(i) For polynomial vector fields the two methods coincide bit-for-bit
(Lemma~\ref{lem:ablation}); no polynomial ARCH benchmark can separate
expansion bases. (ii) For entire nonlinearities ($\sin$, $\cos$, $\exp$)
the Jacobi--Anger expansion bounds the advantage by the constant
$2^{N/(N+1)}\le2$. (iii) For singularities \emph{on} the real axis
(e.g.\ $x^{-3/2}$ gravity terms approached radially) both bases converge
geometrically for every $w$ and neither has a cap; the Chebyshev
advantage is only asymptotic in the degree, and at the degrees used
here the two certified walls differ by a factor of order one.
\end{proposition}

\begin{proof}
(i) is Lemma~\ref{lem:ablation}(b): for polynomial dynamics every
propagation step is a ring operation executed by the same floating-point
instructions in both methods, and the abstraction
\eqref{eq:abstraction} is never invoked, so the certified sets agree
bit for bit. (ii) On $x=m+wt$, $t\in[-1,1]$, combine the angle
addition $\sin(m+wt)=\sin m\cos(wt)+\cos m\sin(wt)$ with the
Jacobi--Anger expansions
$\cos(wt)=J_0(w)+2\sum_{j\ge1}(-1)^{j}J_{2j}(w)\,T_{2j}(t)$ and
$\sin(wt)=2\sum_{j\ge0}(-1)^{j}J_{2j+1}(w)\,T_{2j+1}(t)$: the
Chebyshev coefficients of $\sin$ on the interval are Bessel values
with $|a_j|\le2|J_j(w)|\le 2\,(w/2)^{j}/j!$. The Taylor coefficients
at $m$ are $|c_j|=|\sin^{(j)}(m)|\,w^{j}/j!\le w^{j}/j!$, and since
$\sin^{2}+\cos^{2}=1$ at least one of any two consecutive $j$
attains a fraction $\ge1/\sqrt2$ of this. Both tails are dominated
by their leading terms (the term ratios are $w/(2j)$ and $w/j$), so
the certified radii scale as $r_T\approx w^{N+1}/(N+1)!$ and
$r_C\approx2\,(w/2)^{N+1}/(N+1)!$. Setting each equal to $\tol$ and
dividing, the certified walls obey
$\ws_C/\ws_T=2\cdot2^{-1/(N+1)}=2^{N/(N+1)}\le2$. The same argument
runs for $\cos$, and for $\exp$ with modified Bessel coefficients.
(iii) If the singularity sits on the real axis at distance $\delta$
beyond the endpoint, then it moves \emph{with} the enclosure: its
distance from the center is $\dsing=w+\delta$, so the Taylor rate
$(w/(w+\delta))^{N}$ is geometric for every $w$ and there is no cap.
The Chebyshev parameter is the Joukowski preimage of the real point
at $\delta$ past the endpoint,
$\rho^{*}=1+\delta/w+\sqrt{(1+\delta/w)^{2}-1}$, also geometric, with
log-rate $\sqrt{2\delta/w}\,(1+o(1))$ against Taylor's $\delta/w$ as
$\delta/w\to0$. Setting each rate to $\tol$ gives walls of order
$N\delta/\log(1/\tol)$ for Taylor and $2\delta N^{2}/\log^{2}(1/\tol)$
for Chebyshev: the ratio $2N/\log(1/\tol)$ grows with the degree but
is $0.6$ at $N=4$, $\tol=10^{-6}$, and the shipped Chebyshev magnitude
bounds lose further constants near an endpoint singularity;
Section~\ref{sec:limits} measures $1.02$.
\end{proof}

Section~\ref{sec:limits} verifies each of these as a prediction: a
criterion that forecasts its own null cases is a testable theory.

\subsection{The criterion in action: output sets}\label{sec:gallery}

Scalar walls are the cause; output sets are what a verification tool
actually sees. Two constructions put the dichotomy on display before
any dynamics enter.

\begin{figure}[t]
\centering
\includegraphics[width=\columnwidth]{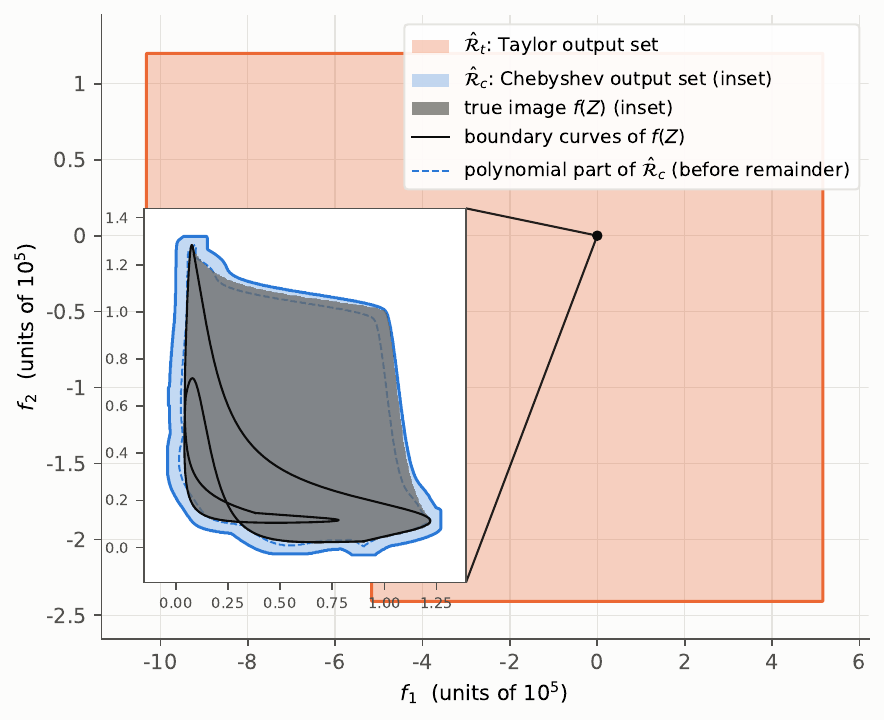}
\caption{The rational map $f(x)=(R(x_1)-0.4x_2,\;R(x_2)+0.4x_1)$,
$R(t)=1/(1+25t^2)$, on the zonotope $Z$ with center $0$ and
generators $(0.45,0.30)^{\top}$, $(0.30,-0.35)^{\top}$; the Taylor
per-coordinate convergence region $|x_i|<1/5$ covers only the middle
of $Z$. Both output sets at total degree 10 in the same
PZ-plus-remainder representation. Taylor: $r_T\approx(5.1\times10^{5},
1.2\times10^{5})$, area $4.6\times10^{11}\times$ the true image.
Chebyshev (inset, $\times\!\sim\!10^{6}$ zoom): $r_C\approx(0.053,0.034)$,
area $1.17\times$; sub-pixel at the outer scale.}
\label{fig:runge2d}
\end{figure}

\begin{figure}[t]
\centering
\includegraphics[width=\columnwidth]{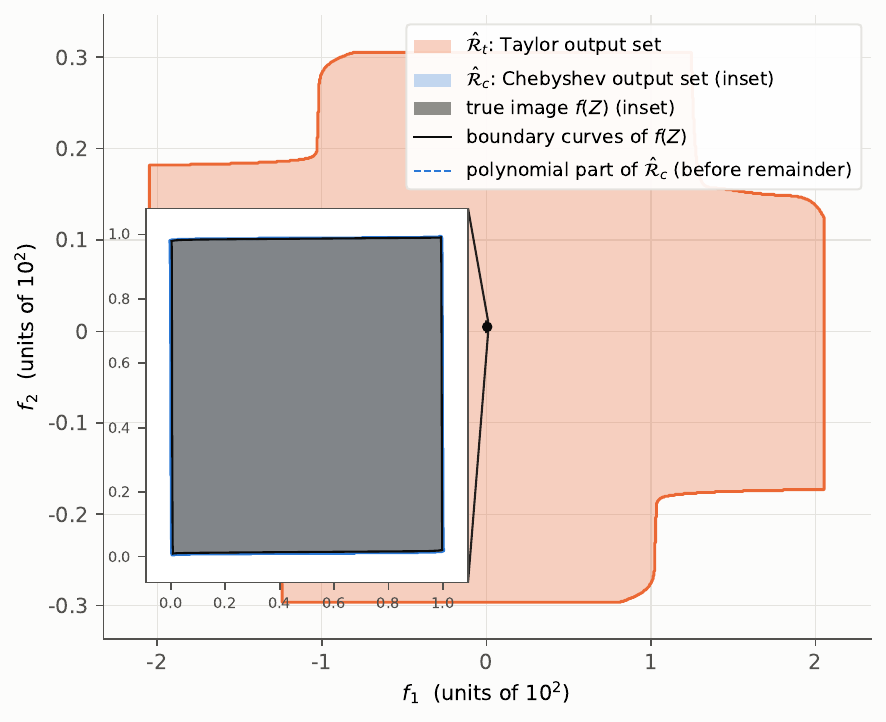}
\caption{The same experiment through a sigmoid layer $f(x)=\sigma(Wx)$ on
a wide logit zonotope (pre-activations span $\approx[-5,5]$, wider
than the pole height $\pi$). The Taylor output set spans $\sim10^{2}$
per axis;
the true image and the Chebyshev output set lie inside $(0,1)^2$ (inset),
the Chebyshev boundary visually indistinguishable from the truth.}
\label{fig:sigmoid2d}
\end{figure}

Figures~\ref{fig:runge2d} and~\ref{fig:sigmoid2d} instantiate
Theorem~\ref{thm:dichotomy} at the set level: one rational map, one
sigmoid layer, identical degree and representation. The areas differ from
the truth by factors $4.6\times10^{11}$ (Taylor) versus $1.17$
(Chebyshev), and $\sim10^{4}$ versus indistinguishable.


\section{The drop-in abstraction and a sound softmax transformer}
\label{sec:method}

This section presents the shipped abstraction: closed-form sound
remainder bounds, their proofs, and the softmax transformer built
from them.

\paragraph{How a set crosses a nonlinearity.} Concretely, let the
current enclosure be a set $X$ and let the step map apply a scalar
primitive $g$ (a $\tanh$, a sigmoid, a rational term) to an inner
expression $u(X)$. The pipeline first bounds the argument,
$u(X)\subseteq[m-w,\,m+w]$. The Chebyshev step then approximates
$g$ \emph{on that whole range}: it takes the degree-$N$ Chebyshev
interpolant $q_{N}$ of $g$ on $[m-w,\,m+w]$ together with a
rigorous remainder radius
$r\ge\sup_{|x-m|\le w}|g(x)-q_{N}(x)|$, and replaces
\begin{equation}\label{eq:chebstep}
g\big(u(X)\big)\;\subseteq\;q_{N}\big(u(X)\big)\oplus[-r,\,r].
\end{equation}
The polynomial image $q_{N}(u(X))$ is computed \emph{exactly} in
the set representation (sums and products of sets;
Section~\ref{sec:pz}), so the entire over-approximation cost of the
crossing is the single number $r$, and $r$ is small exactly when one
degree-$N$ polynomial can track $g$ \emph{uniformly} over the range
the set actually occupies. A Taylor step has the same shape with
$q_{N}$ replaced by the Taylor polynomial at the center $m$: the
anchor is one point, and the same $r$ must absorb everything the
pointwise expansion misses across the rest of the range. At equal
degree on the Runge function of Figure~\ref{fig:runge1d}, that
difference is seven orders of magnitude in $r$. Section~\ref{sec:abstraction}
formalizes this interface for both bases, and
Section~\ref{sec:theory} predicts $r$ from the geometry of $g$.

\subsection{Chebyshev coefficients with closed-form sound remainders}
\label{sec:remainders}

The abstraction ships the degree-$N$ Chebyshev interpolant of $g$ on
$[m-w,m+w]$ (coefficients by a cosine transform, the same $O(N^2)$ as the
Taylor recurrences used by existing tools) together with a remainder
radius that is the \emph{minimum of a family of independently sound
closed-form bounds}. That the pointwise minimum of sound bounds is sound is a
standard device of interval analysis \citep{moore2009interval};
what we have not found elsewhere is a reachability method that
ships its per-primitive remainders this way, for both bases alike
(recent work pursues tighter Taylor remainders by automatic
derivation instead \citep{streeter2023autobound}). The families:
\begin{itemize}
\item \emph{Bernstein-ellipse family} \eqref{eq:cheb-rate} over a grid of
$\rho$, with $M(\rho)$ bounded on a \emph{shifted} complex box
$[m-a_\rho,\,m+a_\rho]\times[-b_\rho,b_\rho]$. The shift is essential:
symmetric-about-zero boxes are blind to the distance $|m|$ from the
switching surface and would erase the surface-distance dependence
of Corollary~\ref{cor:valley}; they would also,
conservatively, overstate our advantage, so the sharper bound is
applied to the Taylor baseline as well.
\item \emph{Elliptic-distance bound} for rational poles: for
$g=1/(1+x^2)$, $|1+z^2|=|z-i||z+i|$ is lower-bounded on $E_\rho$ through
the factorization of the Joukowski map, giving
$M(\rho)\le \big(w\,[(\rho_0-\rho)(\rho-1/\rho_0)/(2\rho)]\big)^{-2}$
with $\rho_0$ the elliptic radius of the pole (new, and mid-aware).
\item \emph{Cauchy tail and exact real Lagrange bounds} for
real-axis singularities ($1/x$, $x^{-3/2}$).
\item \emph{Interval-Lagrange members}, for every primitive whose
$(N{+}1)$-st derivative admits a closed-form supremum, including the
gates through the pole series of the logistic function
(Proposition~\ref{prop:tayfam}(iii)): the classical derivative-form
Lagrange bound joins the family of \emph{both} bases. On Chebyshev nodes the nodal polynomial is minimal among
monic polynomials, $\sup_X\prod_i|x-x_i| = w^{N+1}2^{-N}$, so the
\emph{same} derivative information certifies a factor $2^{N}$
tighter than at a Taylor anchor:
\[
\sup_X|g-q_N| \;\le\;
\frac{\sup_X|g^{(N+1)}|\;w^{N+1}}{(N{+}1)!\,2^{N}} .
\]
This matters at small $w$, where derivative-form bounds beat the
analytic families; the factor $2^{N}$ is free.
\end{itemize}
\begin{figure}[t]
\centering
\includegraphics[width=\columnwidth]{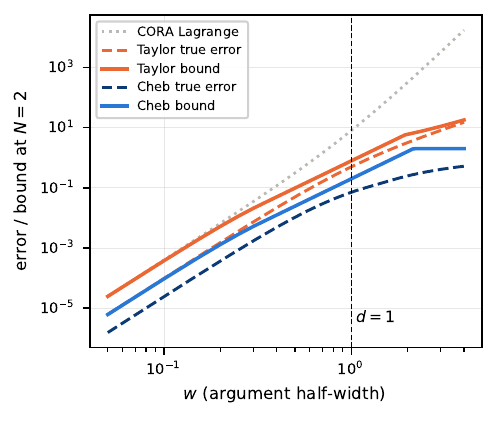}
\caption{Shipped remainder bounds at $N=2$ for $1/(1+x^{2})$ on
$[-w,w]$: the family minimum of each basis against the single-recipe
interval-Lagrange remainder of conservative polynomialization
\citep{althoff2013reachability}; true errors dashed.}
\label{fig:family}
\end{figure}

Figure~\ref{fig:family} grades the discipline scalar-for-scalar
against single-recipe practice. The reference is the
interval-Lagrange remainder used by conservative polynomialization
\citep{althoff2013reachability}: the $(N{+}1)$-st derivative,
bounded by interval arithmetic over the argument range, times
$w^{N+1}/(N{+}1)!$. At degree $N=2$ on $1/(1+x^{2})$ the Taylor
family matches that recipe where it is tight and replaces it where
it is not: at $w=0.1$ the family is already $6\%$ tighter, and past
the pole distance the single recipe overestimates the true error by
two orders of magnitude and keeps growing, while the family's
trivial cap stays within a factor $1.2$ of the truth, $967\times$
tighter at $w=4$. The Chebyshev family sits a further factor below,
$2^{N}$ of it from the node-Lagrange member, and from moderate
widths on the Chebyshev \emph{bound} is smaller than the Taylor
\emph{true} error. The gap has a mathematical source: the family
bounds the derivative supremum through the exact partial-fraction
identity of Proposition~\ref{prop:tayfam} and so sees the zeros of
$g^{(N+1)}$ that any uniform estimate must ignore, and because
sound bounds are closed under pointwise minima, adopting a
competitor's bound into the family can only tighten it.

The rest of this section proves every bound in the family. A
\emph{sound} bound is any $r$ with $\sup_{X}|g-q|\le r$ for the
shipped polynomial $q$; if $r_{1},\dots,r_{J}$ are each sound then
$\min_{j}r_{j}$ is sound, so the family may be evaluated in whole
and the pointwise minimum kept.

\subsection{Magnitude bounds on boxes and ellipses}\label{sec:magnitudes}

The Chebyshev bounds below need, for each primitive, a closed-form
number $M$ with $|g(z)|\le M$ on a complex box
$B=[x_{\mathrm{lo}},x_{\mathrm{hi}}]\times[-Y,Y]$.

\begin{lemma}[Box magnitudes]\label{lem:boxmag}
Let $z=x+iy$ with $x\in[x_{\mathrm{lo}},x_{\mathrm{hi}}]$,
$|y|\le Y$.
\begin{enumerate}
\item[(i)] For $Y<\pi/2$,
\begin{align*}
|\tanh z|^{2}
&=\frac{\sinh^{2}x+\sin^{2}y}{\sinh^{2}x+\cos^{2}y}\\
&\le\max\big(G(s_{\min}),\,G(s_{\max})\big),
\end{align*}
with $G(s)=(s+\sin^{2}Y)/(s+\cos^{2}Y)$, where $s_{\min}$ and
$s_{\max}$ are the minimum and maximum of
$\sinh^{2}x$ over the real range ($s_{\min}=0$ if
$0\in[x_{\mathrm{lo}},x_{\mathrm{hi}}]$, otherwise the smaller
endpoint value).
\item[(ii)] $|\sigma(z)|\le\tfrac12\big(1+M_{\tanh}(B/2)\big)$ for
$Y<\pi$, where $M_{\tanh}(B/2)$ is the bound of (i) on the halved
box.
\item[(iii)] $|\sin z|,|\cos z|\le\cosh Y$ and
$|e^{z}|=e^{x}\le e^{x_{\mathrm{hi}}}$.
\item[(iv)] For $Y<\pi$,
$|\mathrm{softplus}(z)|\le
\mathrm{softplus}(x_{\mathrm{hi}})
+\big((x_{\mathrm{hi}}-x_{\mathrm{lo}})+Y\big)\,M_{\sigma}(B)$,
with $M_{\sigma}(B)$ the bound of (ii).
\end{enumerate}
\end{lemma}

\begin{proof}
(i) From $\tanh(x+iy)=\dfrac{\sinh 2x+i\sin 2y}{\cosh 2x+\cos 2y}$,
\[
|\tanh(x+iy)|^{2}
=\frac{\sinh^{2}2x+\sin^{2}2y}{(\cosh 2x+\cos 2y)^{2}}
=\frac{\sinh^{2}x+\sin^{2}y}{\sinh^{2}x+\cos^{2}y},
\]
where the second equality follows from the double-angle identities
$\sinh^{2}2x=4\sinh^{2}x(1+\sinh^{2}x)$,
$\sin^{2}2y=4\sin^{2}y(1-\sin^{2}y)$,
$\cosh 2x+\cos 2y=2(\sinh^{2}x+\cos^{2}y)$. For fixed $x$ the
expression increases in $\sin^{2}y$, so the supremum over
$|y|\le Y<\pi/2$ sits at $y=\pm Y$. As a function of
$s=\sinh^{2}x\ge0$,
$G(s)=(s+\sin^{2}Y)/(s+\cos^{2}Y)$ has derivative of constant sign
$(\cos^{2}Y-\sin^{2}Y)/(s+\cos^{2}Y)^{2}$, hence is monotone in $s$
and takes its maximum at $s_{\min}$ or $s_{\max}$. If the box meets
the pole line ($Y\ge\pi/2$) the bound is $+\infty$, which is
sound (no certificate).
(ii) $\sigma(z)=\tfrac12(1+\tanh(z/2))$ and the triangle inequality.
(iii) $|\sin(x+iy)|^{2}=\sin^{2}x+\sinh^{2}y\le\cosh^{2}y$, same for
cosine; $|e^{z}|=e^{x}$.
(iv) $\mathrm{softplus}(z)=\log(1+e^{z})$ (principal branch) is
analytic on the strip $|\mathrm{Im}\,z|<\pi$: its singularities,
logarithmic branch points where $1+e^{z}=0$, sit at
$z=i\pi(2n{+}1)$ and are excluded, and $1+e^{z}$ never crosses the
negative real axis there, so the branch is single-valued. Its
derivative is $\sigma$. Join $x_{\mathrm{hi}}$ (real) to
$z=x+iy$ by the horizontal segment from $x_{\mathrm{hi}}$ to $x$
followed by the vertical segment from $x$ to $z$; both segments lie
in $B$ and their total length is at most
$(x_{\mathrm{hi}}-x_{\mathrm{lo}})+Y$. Then
$\mathrm{softplus}(z)=\mathrm{softplus}(x_{\mathrm{hi}})
+\int_{\gamma}\sigma$, and the claim follows from
$\mathrm{softplus}(x_{\mathrm{hi}})\ge0$ and the ML inequality.
\end{proof}

Let $E_{\rho}$ be the closed Bernstein ellipse for $X$: the image of
the disk $|v|\le\rho$ under $z=m+w\,J(v)$, $J(v)=(v+1/v)/2$. Its
semi-axes are $a=w(\rho+1/\rho)/2$ and $b=w(\rho-1/\rho)/2$, so
$E_{\rho}\subseteq[m-a,\,m+a]\times[-b,\,b]$ and every bound of
Lemma~\ref{lem:boxmag} applies with
$x_{\mathrm{lo}}=m-a$, $x_{\mathrm{hi}}=m+a$, $Y=b$. The rational
primitives get sharper, singularity-aware bounds.

\begin{lemma}[Rational magnitudes on $E_\rho$]\label{lem:ratmag}
Let $z\in E_{\rho}$.
\begin{enumerate}
\item[(i)] If $x_{\mathrm{lo}}=m-a>0$ then
$\min_{E_{\rho}}|z|=x_{\mathrm{lo}}$, hence
$|z^{p}|\le x_{\mathrm{lo}}^{\,p}$ for every $p<0$.
\item[(ii)] $|1+z^{2}|\ge 1-b^{2}$ whenever $b<1$.
\item[(iii)] Let $z_{0}\notin E_{\rho}$ have elliptic radius
$\rho_{0}>\rho$, i.e.\ $z_{0}=m+wJ(v_{1})$ with $|v_{1}|=\rho_{0}$.
Then
\[
|z-z_{0}|\;\ge\;
D(\rho,\rho_0)
=w\,\frac{(\rho_{0}-\rho)(\rho-1/\rho_{0})}{2\rho}.
\]
For the conjugate pole pair $\pm i$ of $1/(1+z^{2})$ (equal
$\rho_{0}$ by symmetry when $m$ is real),
$|1+z^{2}|=|z-i|\,|z+i|\ge D(\rho,\rho_0)^{2}$.
\end{enumerate}
\end{lemma}

\begin{proof}
(i) $E_{\rho}$ is convex and symmetric about the real axis. For any
$z=x+iy\in E_{\rho}$, its conjugate is in $E_{\rho}$, and by
convexity so is $x=\tfrac12(z+\bar z)$, with $|x|\le|z|$. Hence the
minimum modulus over $E_{\rho}$ is attained on the real segment
$[m-a,\,m+a]$, where it equals $x_{\mathrm{lo}}$ when
$x_{\mathrm{lo}}>0$.
(ii) With $z=x+iy$,
\begin{align*}
|1+z^{2}|^{2}-(1-y^{2})^{2}
&=(1+x^{2}-y^{2})^{2}+4x^{2}y^{2}-(1-y^{2})^{2}\\
&=x^{2}\,(x^{2}+2+2y^{2})\;\ge\;0 ,
\end{align*}
so $|1+z^{2}|\ge 1-y^{2}\ge 1-b^{2}$ for $|y|\le b<1$.
(iii) The quantity needed is
$\sup_{E_{\rho}}|1/(1+z^{2})|$ over the closed elliptical region;
since the function is analytic there (the poles lie outside), the
maximum-modulus principle puts the supremum on the boundary curve,
so it suffices to bound $|z-z_{0}|$ from below for boundary points
$z=m+wJ(v)$ with $|v|=\rho$ exactly. There
\[
z-z_{0}=\frac{w}{2}\Big(v+\frac1v-v_{1}-\frac1{v_{1}}\Big)
=\frac{w}{2}\,\frac{(v-v_{1})(v v_{1}-1)}{v\,v_{1}} ,
\]
and with $|v-v_{1}|\ge|v_{1}|-|v|=\rho_{0}-\rho$ and
$|v v_{1}-1|\ge|v||v_{1}|-1=\rho\rho_{0}-1$,
\[
|z-z_{0}|\;\ge\;\frac{w}{2}\,
\frac{(\rho_{0}-\rho)\,(\rho\rho_{0}-1)}{\rho\,\rho_{0}}
=w\,\frac{(\rho_{0}-\rho)(\rho-1/\rho_{0})}{2\rho}.
\]
This member enters the family only when it is positive, i.e.\
$\rho<\rho_{0}$ and $\rho\rho_{0}>1$, which is exactly when the
ellipse excludes the pole.
\end{proof}

\subsection{Soundness of the shipped families}\label{sec:famproofs}

\begin{proposition}[Chebyshev remainders]\label{prop:chebfam}
Let $q$ be the degree-$N$ Chebyshev interpolant of $g$ on $X$ and
let $c_{0},\dots,c_{N}$ be its coefficients. Each of the following
is a sound bound on $\sup_{X}|g-q|$:
\begin{enumerate}
\item[(i)] for every $\rho>1$ such that $g$ is analytic on
$E_{\rho}$ and $|g|\le M(\rho)$ there,
$\dfrac{4M(\rho)}{(\rho-1)\,\rho^{N}}$, with $M(\rho)$ any bound
from Lemma~\ref{lem:boxmag} applied to the bounding box of
$E_{\rho}$, or from Lemma~\ref{lem:ratmag} for the rational
primitives;
\item[(ii)] $M_{\mathrm{real}}+\sum_{j\le N}|c_{j}|$, where
$M_{\mathrm{real}}=\sup_{X}|g|$ is evaluated from the exact
monotone or unimodal range of the primitive;
\item[(iii)] (Chebyshev-node Lagrange)
$\dfrac{w^{N+1}}{(N{+}1)!\,2^{N}}\,\sup_{X}|g^{(N+1)}|$ whenever
the derivative supremum has a closed form (the same list as
Proposition~\ref{prop:tayfam}(iii)).
\end{enumerate}
The family takes the minimum of (i) over a fixed grid of $\rho$
values, (ii), and (iii).
\end{proposition}

\begin{proof}
(i) is the Chebyshev interpolation bound of
\citet[Thm.~8.2]{trefethen2019atap}: if $g$ is analytic in
$E_{\rho}$ with $|g|\le M(\rho)$, then
$\sup_X|g-q|\le 4M(\rho)\rho^{-N}/(\rho-1)$ (stated there for
Chebyshev points of the second kind; the aliasing argument gives the
same constant for the first-kind points used here). The hypotheses hold
because each $M(\rho)$ of Lemmas~\ref{lem:boxmag}
and~\ref{lem:ratmag} is finite only when the corresponding box or
ellipse excludes every singularity of the primitive: the tanh and
sigmoid bounds are finite only below the pole line, the softplus
bound only inside the strip of analyticity, the leftmost-vertex
bound only when the ellipse stays right of the origin, and the
elliptic-distance bound only when the pole's elliptic radius
exceeds $\rho$. Analyticity on $E_\rho$ is therefore implied by
finiteness of the bound, and an infinite family member is simply
never the minimum. (ii) is the triangle inequality with
$|T_{j}|\le1$ on $X$:
$|g-q|\le|g|+\sum_j|c_{j}||T_{j}|\le
M_{\mathrm{real}}+\sum_j|c_{j}|$. (iii) The interpolation error at
nodes $x_{0},\dots,x_{N}$ is
$g(x)-q(x)=\frac{g^{(N+1)}(\xi)}{(N+1)!}\prod_{i}(x-x_{i})$ for
some $\xi\in X$, and for Chebyshev points of the first kind the
nodal polynomial is the rescaled $T_{N+1}$, minimal among monic
polynomials: $\sup_{X}\prod_{i}|x-x_{i}|=w^{N+1}2^{-N}$. The same
derivative suprema as in Proposition~\ref{prop:tayfam}(iii) apply,
and the factor $2^{N}$ over the Taylor anchor is free.
\end{proof}

\begin{proposition}[Taylor remainders]\label{prop:tayfam}
Let $T_{N}$ be the degree-$N$ Taylor polynomial of $g$ at $m$ with
coefficients $b_{0},\dots,b_{N}$. Each of the following is a sound
bound on $\sup_{X}|g-T_{N}|$:
\begin{enumerate}
\item[(i)] (coefficient tail) for every $r>1$ such that $g$ is
analytic on the disk $|z-m|\le wr$ and bounded there by $M_{r}$,
$\dfrac{M_{r}}{r^{N}\,(r-1)}$;
\item[(ii)] (Lagrange--Cauchy band) for every $s>0$ such that $g$
is analytic and bounded by $M_{s}$ on the band
$\{|\mathrm{Re}\,z-m|\le w+s,\ |\mathrm{Im}\,z|\le s\}$,
$M_{s}\,(w/s)^{N+1}$;
\item[(iii)] (exact Lagrange) $\dfrac{w^{N+1}}{(N+1)!}\,
\sup_{X}|g^{(N+1)}|$ whenever the derivative supremum has a closed
form ($\sin$, $\cos$: $1$; $e^{x}$: $e^{m+w}$; $x^{p}$ with $p<0$
on $m-w>0$: the left endpoint value, since $|g^{(N+1)}|$ is
decreasing; $1/(1+x^{2})$: from the exact identity
$g^{(j)}(x)=(-1)^{j}j!\,\mathrm{Im}\big[(x-i)^{-(j+1)}\big]$,
evaluated on a Lipschitz-padded grid, capped by the uniform bound
$(N{+}1)!\,(1+x_{*}^{2})^{-(N+2)/2}$ with $x_{*}$ the point of $X$
nearest $0$; $\sigma$, $\tanh$ and softplus: from the pole series
$\sigma^{(j)}(x)=(-1)^{j}j!\,2\,\mathrm{Re}\sum_{n\ge0}
\big(x-i\pi(2n{+}1)\big)^{-(j+1)}$ with
$\tanh^{(j)}(x)=2^{j+1}\sigma^{(j)}(2x)$ and
$\mathrm{softplus}^{(j)}=\sigma^{(j-1)}$, evaluated the same way with
the truncation tail $\pi^{-(j+1)}(2K{-}1)^{-j}/j$ after $K$ terms and
capped by the uniform bound $j!\,2\pi^{-(j+1)}(1+1/(2j))$);
\item[(iv)] (trivial) $M_{\mathrm{real}}+\sum_{j\le N}|b_{j}|w^{j}$.
\end{enumerate}
The family takes the minimum over grids of $r$ and $s$; the
magnitude $M_{r}$ is evaluated with Lemma~\ref{lem:boxmag} on the
bounding box of the disk, and the rational $M_{r}$ uses the
per-pole distance bound
$|1+z^{2}|\ge\big(\sqrt{1+m^{2}}-wr\big)^{2}$ on the disk.
\end{proposition}

\begin{proof}
(i) By the Cauchy estimates on the disk of radius $R=wr$ centered
at $m$, the Taylor coefficients of $g$ at $m$ satisfy
$|a_{j}|\le M_{r}/R^{j}$. Hence on $X$,
\[
|g-T_{N}|\le\sum_{j\ge N+1}|a_{j}|\,w^{j}
\le M_{r}\sum_{j\ge N+1}r^{-j}
=\frac{M_{r}}{r^{N}(r-1)} .
\]
(ii) The real Lagrange form gives
$|g-T_{N}|\le\frac{w^{N+1}}{(N+1)!}\sup_{\xi\in X}|g^{(N+1)}(\xi)|$.
For each real $\xi\in X$ the disk $|z-\xi|\le s$ lies in the stated
band, so the Cauchy derivative estimate gives
$|g^{(N+1)}(\xi)|\le (N+1)!\,M_{s}/s^{N+1}$, and the factorials
cancel. (iii) is the Lagrange form with the exact derivative
supremum; for $x^{p}$, $|g^{(N+1)}(x)|=|p(p-1)\cdots(p-N)|\,
x^{\,p-N-1}$ is decreasing on $x>0$, so the supremum sits at
$m-w$. For $1/(1+x^{2})=\frac{1}{2i}\big((x-i)^{-1}-(x+i)^{-1}\big)$,
differentiating the partial fractions and using
$(x+i)^{-(j+1)}=\overline{(x-i)^{-(j+1)}}$ for real $x$ gives the
identity $g^{(j)}(x)=(-1)^{j}j!\,
\mathrm{Im}\big[(x-i)^{-(j+1)}\big]$, hence the uniform bound
$|g^{(j)}|\le j!\,(1+x^{2})^{-(j+1)/2}$ from
$|x\mp i|=\sqrt{1+x^{2}}$; the supremum of the exact identity over
$X$ is bounded by its maximum on a uniform grid of spacing $\delta$
plus the Lipschitz term $\sup|g^{(j+1)}|\,\delta/2$, which is
sound and, unlike the uniform bound, sees the zeros of the
derivative. For the gates, $\sigma(z)=1/(1+e^{-z})$ has simple poles
at $z_n=i\pi(2n{+}1)$, $n\in\mathbb{Z}$, each with residue $1$
(the denominator has derivative $-e^{-z_n}=1$ there), and is bounded
on the strips between them, so its Mittag-Leffler expansion is
$\sigma(z)=\tfrac12+\sum_{n\ge0}2z/\big(z^{2}+\pi^{2}(2n{+}1)^{2}\big)
=\tfrac12+\sum_{n\ge0}\big[(z-z_n)^{-1}+(z+z_n)^{-1}\big]$;
differentiating termwise for $j\ge1$ and pairing the conjugate poles
gives the stated series, the tail after $K$ terms is bounded by
comparison with $\int_{K-1}^{\infty}(2t{+}1)^{-(j+1)}\,dt$, and the
uniform bound uses $|x-z_n|\ge\pi(2n{+}1)$ on the real line. The
identities $\tanh(x)=2\sigma(2x)-1$ and $\mathrm{softplus}'=\sigma$
transfer the bound. For the rational disk bound in the last claim: each pole
$\pm i$ has $|z\mp i|\ge|m\mp i|-|z-m|\ge\sqrt{1+m^{2}}-wr$,
and $|1+z^{2}|$ is the product of the two distances. (iv) as in
Proposition~\ref{prop:chebfam}(ii) with $|(x-m)^{j}|\le w^{j}$.
\end{proof}

\subsection{A sound softmax abstract transformer}\label{sec:softmax}

The family above covers the scalar nonlinearities a controller
contains; what remains is multivariate. Softmax has no sound
set-level treatment in the literature, and we obtain one by
certifying the numerically stable log-sum-exp
decomposition,
\[
a_i \;=\; \sigma\big(-\mathrm{LSE}_i\big),\qquad
\mathrm{LSE}_i \;=\; d_{1}+\mathrm{softplus}(d_{2}-d_{1}),
\]
with $d_{1},d_{2}$ the exact PZ differences of the competing logits.
Every factor is univariate and strip-analytic on
$|\mathrm{Im}\,z|<\pi$ ($\sigma$ with simple poles at $\pm i\pi$,
softplus with logarithmic branch points there), so the machinery of
Section~\ref{sec:remainders} applies; the softplus remainder uses an
anchored-integral bound $M_{\mathrm{sp}}\le
\mathrm{softplus}(x_{\mathrm{hi}})+\ell\,M_{\sigma}$ along a path of
length $\ell$ inside the box (new; proved as
Lemma~\ref{lem:boxmag}(iv)). Why this
decomposition, and not the obvious one: composing
$a_i=1/(1+\sum_{j\ne i}e^{s_j-s_i})$ from exponential and reciprocal
primitives fails structurally: the conservative range of the
denominator crosses zero on wide logit sets and the certificate
collapses.

\begin{theorem}[Sound softmax transformer]\label{thm:softmax}
The transformer above returns, for every logit PZ, componentwise
enclosures of the softmax image and, because the components are built
on shared symbols, sound lower bounds on the decision margins
$a_1-a_j$. It is validated with zero violations against
$2\times10^{5}$ samples per configuration.
\end{theorem}

\begin{proof}
Soundness composes stage by stage
(Lemma~\ref{lem:ablation}(a)). The logit differences
$d_j=s_j-s_i$ are exact PZ subtractions over the shared symbol
vector; no abstraction error enters. The softplus stage applies
\eqref{eq:abstraction} on the certified interval of its argument; its
anchored-integral remainder bound is valid on every real interval
because $\mathrm{softplus}$ is analytic on the strip
$|\mathrm{Im}\,z|<\pi$ and the bound is constructed from the
monotone integrand on the real line, so the singularities at
$\pm i\pi$ never meet the domain of the bound. The final stage
$a_i=\sigma(-\mathrm{LSE}_i)$ applies the sigmoid abstraction on the
certified range of $\mathrm{LSE}_i$, with the same pole clearance by
construction. True-range clipping is sound because $\sigma$ and
$\mathrm{softplus}$ are monotone: the exact image of a sound input
interval is $[\,g(\mathrm{lo}),g(\mathrm{hi})\,]$, which contains the
image of the true set, and intersecting two sound enclosures is
sound. For the margins, $a_1-a_j$ is again a PZ over the same symbol
vector, so its interval lower bound is a bound on the minimum over
the \emph{joint} range of $(a_1,a_j)$; no independence assumption is
introduced at the subtraction, which is exactly where interval
arithmetic loses the anti-correlation of softmax components. The
final sentence is the measured validation protocol of
Section~\ref{sec:experiments}.
\end{proof}

Margins are the point: the decisions of Section~\ref{sec:radius}
are provable exactly because the subtraction happens on shared
symbols.

\begin{figure*}[t]
\centering
\includegraphics[width=\textwidth]{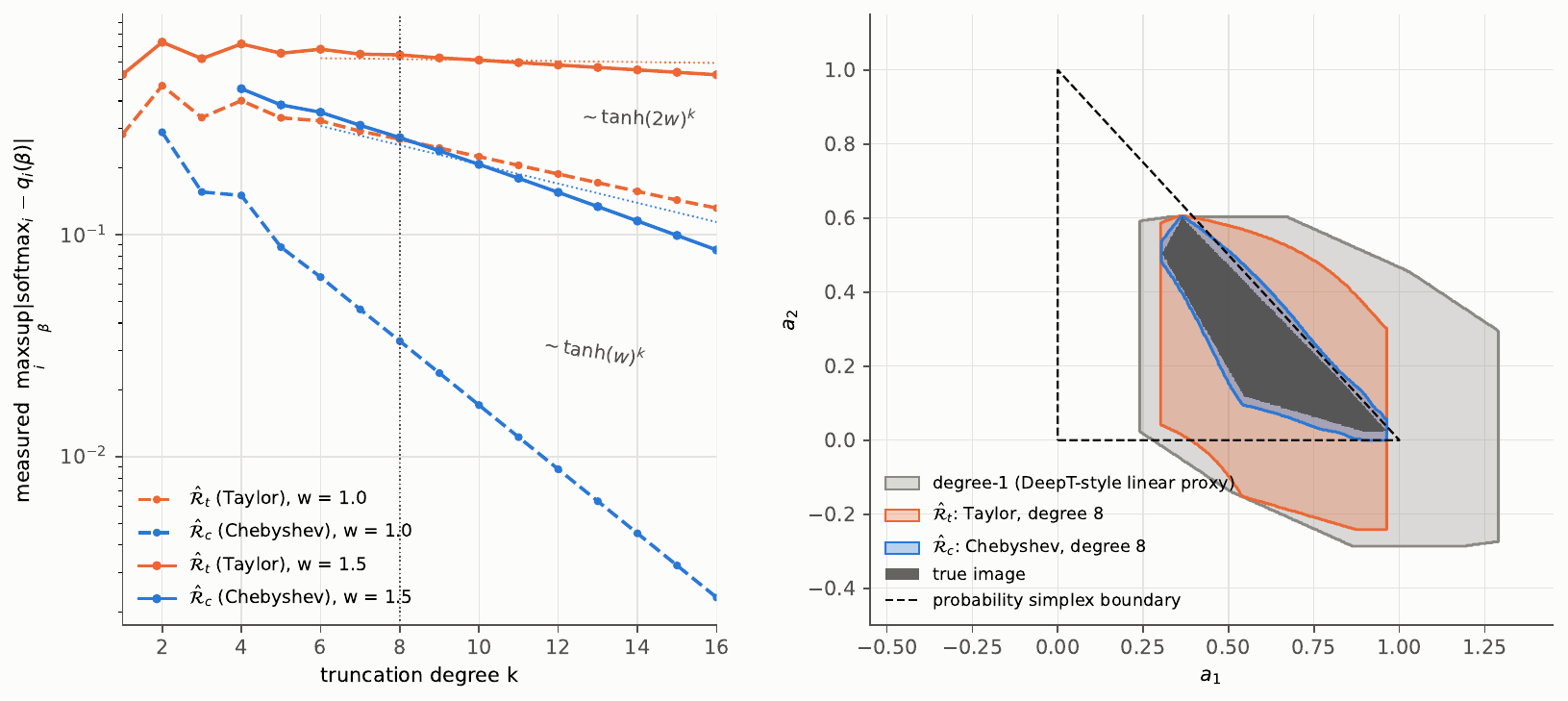}
\caption{Prototype softmax abstraction on a logit zonotope, identical
pipeline. \emph{Left:} end-to-end measured remainder versus degree for
logit half-width $w\in\{1.0,1.5\}$: Taylor $\sim\tanh(2w)^{N}$,
Chebyshev $\sim\tanh(w)^{N}$ (dotted vertical: $N=8$, right panel).
\emph{Right:} output sets in the $(a_1,a_2)$ projection (logit
centers $(1.5,0,-1)$); the dashed triangle is the boundary of the
probability simplex ($a_i\ge0$, $a_1+a_2\le1$), which every true
softmax output lies inside, so certified area outside it is pure
over-approximation. Degree-1 linear relaxation $6.3\times$ the true
area, Taylor $N=8$: $3.6\times$, Chebyshev $N=8$: $1.3\times$.
Rigorous decision-radius numbers in Table~\ref{tab:radius}.}
\label{fig:softmax}
\end{figure*}

Together, Lemma~\ref{lem:boxmag}(iv) and
Proposition~\ref{prop:chebfam} close the two openings in the proof
of Theorem~\ref{thm:softmax}: the softplus stage's
anchored-integral bound is Lemma~\ref{lem:boxmag}(iv), and every
remainder radius it composes with is covered by
Proposition~\ref{prop:chebfam}.


\section{Experiments}\label{sec:experiments}

This section measures certified walls and verdicts under fixed,
pre-registered protocols and compares them with the predictions of
Sections~\ref{sec:theory}--\ref{sec:method}; the predictions that
missed are reported in Section~\ref{sec:limits}.

\paragraph{Protocol.} Benchmarks: the pendulum
amplitude family
(entire-function calibration), a rationally damped oscillator
(\textsc{ratd}: $\dot x_1=x_2$, $\dot x_2=-x_1-x_2/(1+x_1^2)$), a
temperature-gated damping schedule (\textsc{gate}$_\tau$:
$\dot x_2=-x_1-(0.2+1.3\,\sigma(x_1/\tau))x_2$, a smooth gain
schedule), planar Kepler dynamics ($\mu/r^{3}$, real-axis control),
Van der Pol and Duffing (polynomial ties), a $\tanh$-network-controlled
TORA system, and the softmax classifier of
Section~\ref{sec:softmax}. Methods: interval, linear, Taylor-$N$,
Chebyshev-$N$, hybrid, at degrees $2$, $4$, $8$, $16$ and PZ orders
$D\in\{2,3\}$; walls $\ws$ by monotone grid plus bisection. Soundness:
every certified run is checked step-by-step against $2\times10^{5}$
Monte-Carlo samples plus all corners; \emph{total violations: zero} ($6{,}223$ certified runs in the wall
sweeps alone, and every run of the sections below). Predictions were pre-registered;
misses are reported in Section~\ref{sec:limits}.

\subsection{Closed loop: survival, walls, verdicts}
\label{sec:survival}\label{sec:walls}\label{sec:flips}

\begin{figure*}[t]
\centering
\includegraphics[width=\textwidth]{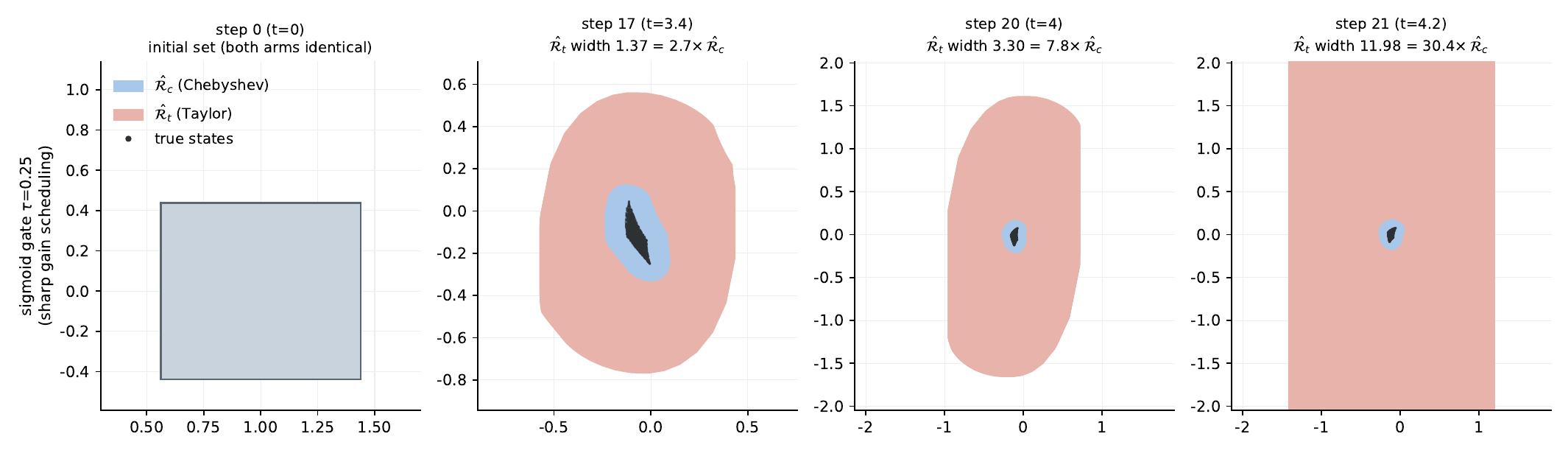}
\caption{Forward propagation on the sigmoid-gate loop ($\tau=0.25$),
identical pipeline (degree $4$, $D=3$), initial half-width between the two
walls; certified sets rendered the polynomial-zonotope way
($16\times16$ domain split, exact re-expansion, bounded residual);
companion to Figure~\ref{fig:ainncs}(c). By $k=17$, $\Rt$ is
$2.7\times\,\Rc$; by $k=20$, $7.8\times$; at $k=21$ it fills the frame
($30\times$) and at $k=22$ the certificate is lost (NaN). $\Rc$ hugs
the Monte-Carlo truth for all 100 steps.}
\label{fig:predeath}
\end{figure*}

The first question is whether a certificate finishes its horizon
at all.
Figure~\ref{fig:ainncs}(c) shows the dichotomy reaching a closed
loop on the rational-damping system; Figure~\ref{fig:predeath} adds the
sigmoid-gate companion.

Both methods receive the same initial set, chosen between the two certified
walls. $\Rt$ inflates by the mechanism of
Theorem~\ref{thm:dichotomy}: each pass of the enclosure through the
gate region multiplies the remainder. It dies at $t=3.4$ (step $17$
of $100$ on the rational-damping loop, step $21$ on the gate loop);
$\Rc$ completes the horizon with its certificate intact.

Survival is binary; the walls measure how much margin separates the
two outcomes.

\begin{table}[t]
\centering\footnotesize\setlength{\tabcolsep}{2pt}
\caption{Certified initial-set walls $\ws$: ratio
Chebyshev/Taylor at degree $4$, $D=3$.}
\label{tab:walls}
\begin{tabular}{lcc}
\toprule
family & scalar type & $\ws$ ratio C/T\\
\midrule
pendulum $A\in[0.5,2.5]$ & $\sin$ (entire) & 1.07--1.16\\
\textsc{ratd} $A\in[0.5,2]$ & $1/(1+x^2)$, poles $\pm i$ & 1.11--1.22\\
\textsc{gate} $\tau\in[0.125,1]$ & $\sigma(x/\tau)$ (poles $\pm i\pi\tau$) & 1.09--1.16\\
TORA + sigmoid/$\tanh$ networks & $\sigma,\tanh$ (poles $\pm i\pi\tau$) & 1.25--1.67\\
ACC + $\tanh$ network & $\tanh$ & 1.05\\
Kepler $\mu/r^{3}$ & $x^{-3/2}$ (real axis) & 1.02\\
Van der Pol / Duffing & polynomial & 1.000 (bit-for-bit)\\
\bottomrule
\end{tabular}
\end{table}

Table~\ref{tab:walls} summarizes the wall matrices. Two observations.
First, polynomial fields tie bit for bit and the real-axis family
ties within $2\%$, as the criterion predicts, while every family
with off-axis singularities separates, entire functions included.
Second, with tight remainders on both sides the closed-loop
separation at the practical degree $N=4$ is compressed into
$1.07$--$1.22$ for the pendulum, rational and gate families, and
reaches $1.34$ and $1.67$ only on the TORA sigmoid and $\tanh$
controllers, whose gates are the sharpest in the table. Both bases saturate against a basis-independent
representation ceiling, analysed with Monte-Carlo attribution in
Section~\ref{sec:limits}, and that ceiling, not the abstraction, is
what keeps the closed-loop ratios far below the scalar factor $2^{N}$
of Section~\ref{sec:method}.

A specification is either proven or not, and the flips below show
that even a $13$--$14\%$ margin is exactly the width of the gap
between \textsc{proven} and \textsc{unknown}.

\begin{table}[t]
\centering\footnotesize\setlength{\tabcolsep}{2pt}
\caption{Safety verdicts at degree $4$, $D=3$: same pipeline,
same tolerance, only the basis differs.}
\label{tab:flips}
\begin{tabular*}{\columnwidth}{@{\extracolsep{\fill}}llcc@{}}
\toprule
system & $\ws$-scale & Taylor & Chebyshev\\
\midrule
\textsc{ratd} & 5.4 & \textsc{proven} & \textsc{proven}\\
\textsc{ratd} & 9.55 & \textsc{unknown}$^{\dagger}$ & \textbf{\textsc{proven}}\\
\textsc{gate}$_{0.25}$ & 4.9 & \textsc{proven} & \textsc{proven}\\
\textsc{gate}$_{0.25}$ & 8.77 & \textsc{unknown}$^{\dagger}$ & \textbf{\textsc{proven}}\\
pendulum & 1.05 & \textsc{unknown} & \textsc{unknown}\\
Kepler & 0.2 & \textsc{proven} & \textsc{proven}\\
\bottomrule
\multicolumn{4}{l}{\footnotesize $^{\dagger}$certificate lost at step 17
(\textsc{ratd}) or 21 (\textsc{gate}), enclosure blow-up;}\\
\multicolumn{4}{l}{\footnotesize \phantom{$^{\dagger}$}invariant and
goal unprovable.}
\end{tabular*}
\end{table}

Table~\ref{tab:flips} translates the walls into verification
outcomes. The specification is an invariant
$\|x\|_\infty\le3.2$ over $100$ steps with a goal
$\|x\|_\infty\le0.5$ at the horizon; every row is truly safe
($5\times10^{4}$ Monte-Carlo trajectories), and interval arithmetic
returns \textsc{unknown} on every row.
On both pole-regime systems there is a regime, truly safe and confirmed
by Monte-Carlo, where the Taylor method answers \textsc{unknown}
because its certificate explodes mid-horizon, and the Chebyshev method
proves both the invariant and the goal. Interval arithmetic proves
nothing anywhere; the pendulum row is a double \textsc{unknown}
(the chosen invariant is tighter than either certified tube at that
scale); the Kepler row ties, as predicted.

\subsection{Certified decision radius for softmax}\label{sec:radius}

The softmax transformer is measured on the decision itself.

\begin{table}[t]
\centering\footnotesize\setlength{\tabcolsep}{3.5pt}
\caption{Certified decision radius on the logit-box prototype
(centers $(1.5,0,-1)$; exact corner oracle $0.750$).}
\label{tab:radius}
\begin{tabular}{lccccccc}
\toprule
 & oracle & interval & linear $N{=}1$
 & \multicolumn{2}{c}{Taylor} & \multicolumn{2}{c}{Chebyshev}\\
 & (exact) & & (T\,/\,C) & $N{=}4$ & $N{=}8$ & $N{=}4$ & $N{=}8$\\
\midrule
radius & 0.750 & 0.456 & 0.576\,/\,0.614 & 0.610 & 0.612
 & \textbf{0.622} & 0.613\\
\bottomrule
\end{tabular}
\end{table}

Table~\ref{tab:radius} is an abstraction-layer stress test: on a
plain logit box the corner oracle is exact and every abstraction sits
below it. With the remainder families of Section~\ref{sec:method}
even the linear ($N{=}1$) abstractions, the relaxation regime of
\citet{shi2020transformers,bonaert2021deept}, certify $77$--$82\%$
of the analytic radius, above interval arithmetic ($61\%$), and the
Chebyshev transformer at degree $1$ already reaches the radius
($0.614$) that Taylor attains only at degree $8$ ($0.612$). Both
bases then plateau near $0.61$--$0.62$ from degree $4$ on, the $D=3$
carrier ceiling ($82\%$ of the truth); the Chebyshev value at degree
$2$, $0.654$, sits above that plateau because a lower abstraction
degree leaves fewer terms to truncate. In network-verification
practice, where every neuron of every layer is abstracted and the
affordable degree per scalar is small, the low-degree columns are
the operative ones. On real trained logits the ordering persists: on
a three-class classifier at $90\%$ accuracy, the certified-abstention
radius, the largest logit uncertainty at which the winning
probability provably stays above $0.5$, averages $1.23\times$ the
Taylor value at equal degree, and Chebyshev at $N{=}4$ exceeds Taylor
at $N{=}8$ by $20\%$.

\subsection{Open-loop neural networks}\label{sec:nnwindow}

Does the basis also pay in open-loop network verification, the
robustness setting of
\citet{katz2017reluplex,singh2019deeppoly,zhang2018crown,%
liu2021algorithms}? The window has walls, and we measured both the
walls and what lies between them. Piecewise-linear networks stay out
of reach on principle (Proposition~\ref{prop:scope}): measured
$1.000$ on the closed-loop AINNCS ReLU controller and $0.85$--$0.89$
on raw ReLU layers, a regime without poles. Small perturbation radii leave
every per-neuron interval in the linear regime, where the bases
agree to eight digits, and large radii make the property itself
false before any abstraction fails. Between those walls the basis
separates as the dichotomy predicts. On random $16$-unit $\tanh$
layers the raw certified width ratio Taylor over Chebyshev grows
monotonically with the pre-activation half-width, $1.02$ at $0.5$,
$3.1$ at $2$, $13$ at $4$ and $45.7$ at $6$ (medians over five
seeds), and composed through depth the ratio settles at $1.3$ by two
to four layers with the Chebyshev enclosure within $1.4$--$1.6\times$
the true output range throughout while Taylor sits at
$1.7$--$2.1\times$. (Production tools add a
per-neuron exact-range fallback; it can mask the single-layer gap,
but it forgets the correlations between neurons, which is precisely
what margins and closed loops consume, so we measure the ablation
raw and treat the fallback as a separate design choice.) On trained
classifiers interval arithmetic, exact per neuron but blind to
correlations, collapses with depth, as the six-layer MNIST network
below shows, while the polynomial's shared symbols hold.

\begin{figure*}[t]
\centering
\includegraphics[width=\textwidth]{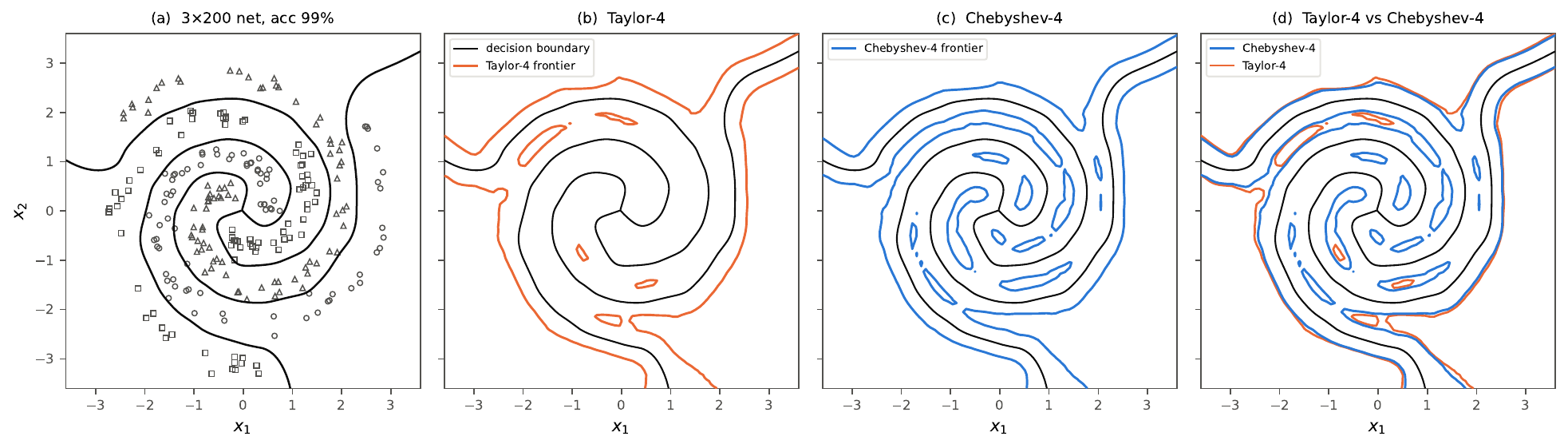}
\caption{Certified frontiers at $\epsilon=0.18$ for a
$3\times200$ $\tanh$ classifier trained on a three-armed spiral
($99\%$ held-out accuracy, $80\times80$ grid). \textbf{(a)}~Held-out
points (marker shape = class) and the network's true decision
boundary in black. \textbf{(b)}~The Taylor-4 certification frontier.
\textbf{(c)}~The Chebyshev-4 frontier at the same degree and cost.
\textbf{(d)}~Both overlaid; the strip between the curves is truly
safe territory that only the basis change recovers.}
\label{fig:nnspiral}
\end{figure*}

\paragraph{Certified frontiers at scale.} We measured full-plane
certification maps on $200$-wide $\tanh$ classifiers trained on a
three-armed spiral ($24{,}000$ training points, weight decay), a
task whose \emph{true} boundary is complex by construction, so
the boundary complexity comes from the task rather than from
memorized noise. The six-hidden-layer network
($99.2\%$ held-out accuracy) is the harder case: at
$\epsilon=0.08$ it leaves $19.1\%$ of the input plane uncertified
under Taylor-4 against $18.0\%$ under Chebyshev-4; at
$\epsilon=0.12$, $35.9\%$ against $31.2\%$; at $\epsilon=0.18$,
$58.0\%$ against $52.0\%$, a strip of $4.7$ to $5.9$ points of the
plane recovered by the change of basis alone at identical degree and
cost. Figure~\ref{fig:nnspiral} draws the mechanism on the
three-hidden-layer sibling ($99.2\%$ accuracy). Each colored curve
is a certification frontier: on its safe side the method proves the
predicted label constant on the entire box $x\pm\epsilon$. The only
inputs truly unsafe at this radius are those whose box straddles
the black decision boundary, an $\epsilon$-collar around it, so
every gap between a frontier and that collar is safe territory the
arithmetic
surrenders: the property holds there, and what fails is the
truncation error, amplified layer by layer by the wrapping effect.
At $\epsilon=0.12$ the two frontiers nearly coincide ($24.7\%$ of
the plane uncertified under Taylor-4 against $23.6\%$); at
$\epsilon=0.18$ Taylor-4 gives up the inner windings outright while
Chebyshev-4 still holds certified islands along them ($46.5\%$
against $39.6\%$, a $6.9$-point strip of truly safe plane recovered
by the basis alone). The separation grows monotonically with radius,
depth, and boundary curvature, the two-regime shape of the closed
loop again. Raising
the Chebyshev degree past four moves these shares by about one
point before flattening, and, unlike the Taylor degree inversion,
more degree never hurts; degree four is used throughout. Order
reduction also becomes real at this width: every $\tanh$ layer
promotes about $200$ fresh pool generators, and forcing the budget
down to $128$ columns costs at most $1.3$ points of plane area
anywhere we measured, because reduction drops only near-massless
columns. Preconditioning the dropped columns in the next layer's
singular frame, ported from the rotated-frame order-reduction
literature \citep{girard2005zonotope,kopetzki2017methods}, never
gains, nil on clean networks and up to $14$ points on an
over-trained variant: our pool columns are born one-hot as
per-neuron remainders, so the axis-aligned hull of the dropped set
is exact and any rotation strictly inflates it, while the interval
transport step the preconditioner is built to save does not exist
in a pipeline that transports generators exactly.

\begin{figure*}[t]
\centering
\includegraphics[width=\textwidth]{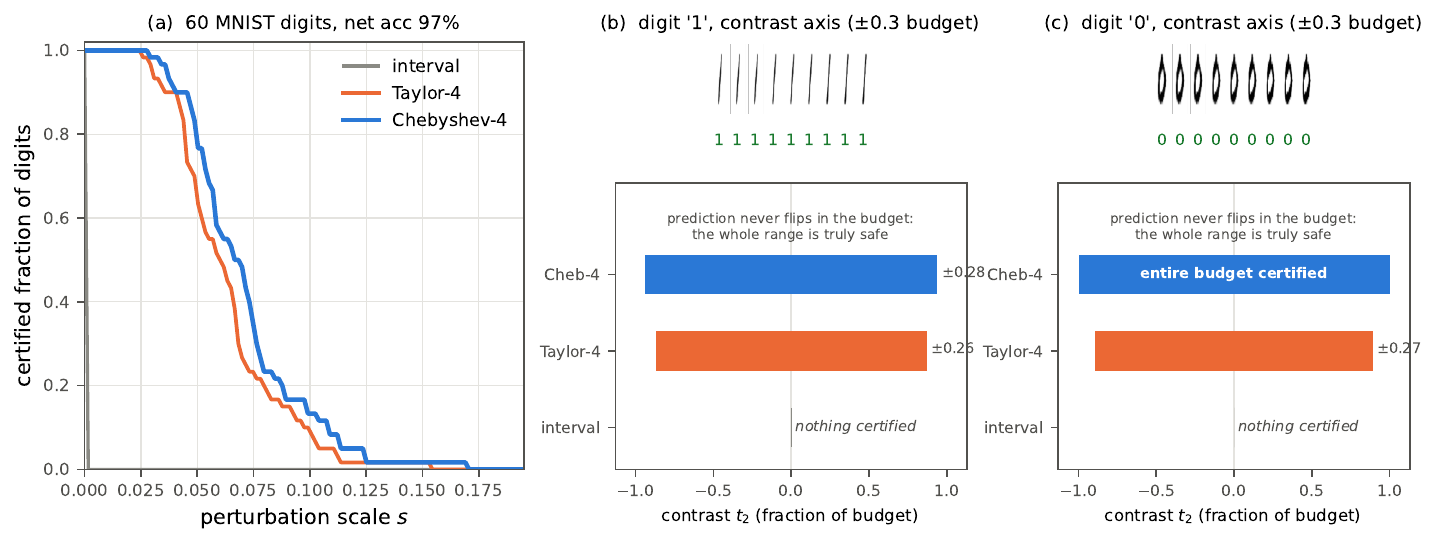}
\caption{MNIST under a certified semantic perturbation.
\textbf{(a)}~Certified coverage of $60$ test digits versus the
fraction $s$ of the brightness-and-contrast budget ($\pm0.30$
each): the Chebyshev curve dominates at every $s$; the interval
curve sits at zero. \textbf{(b,c)}~One digit per panel, contrast
axis. The thumbnail strip shows the digit across the entire budget
with the network's prediction underneath: it never flips, so the
whole range is truly safe. The bars below show how much of that
truth each method can prove. Chebyshev-4 certifies $\pm0.28$ of
the $\pm0.30$ range on the first digit and the entire budget on the
second; Taylor-4 stops at $\pm0.26$ and $\pm0.27$; interval
arithmetic certifies nothing. In practice a bar reads as
a warranty: for any contrast change inside it, the classifier's
answer provably does not move.}
\label{fig:nnmnist}
\end{figure*}

\paragraph{A real image classifier.} The same machinery runs
unchanged on MNIST. The dense degree-$4$ representation carries two
shared symbols across a $784$-dimensional state, so the natural
threat model is a low-dimensional semantic family
\citep{mohapatra2020semantify,balunovic2019deepg} rather than
per-pixel noise, which would demand $784$ symbols and a sparse
representation. The family: every pixel brightened jointly by up to
$\pm0.30$, and contrast scaled jointly by up to $\pm30\%$, one
shared symbol per knob. On a $784\to200{\times}6\to10$ $\tanh$
classifier at $97.1\%$ test accuracy, interval arithmetic certifies
not one of $60$ test digits at any positive fraction of the budget,
the six-layer collapse replayed on real images. Both polynomial
methods survive, and the basis gap persists but is modest here:
Chebyshev-4 certifies a mean $1.12\times$ the Taylor-4 budget
fraction at equal degree and is the larger of the two on $60$ of
$60$ digits. Figure~\ref{fig:nnmnist} shows the deployable reading:
along the contrast axis the prediction never flips anywhere in the
budget, the whole range is truly safe, and Chebyshev-4 certifies the
entire range on $58$ of the $60$ digits, Taylor-4 on $52$, interval
arithmetic on none; the two example digits are ones where the bases
part, Chebyshev proving the full budget or $94\%$ of it where Taylor
stops at $87$--$89\%$.

\subsection{Stress tests and baselines}
\label{sec:splitting}\label{sec:ainncs}\label{sec:anchor}

Every race above ran one set per method, and a practitioner will object
that nobody runs one set. The standard mitigation for a diverging
Taylor abstraction is to split the initial set, the reachability
analogue of branch-and-bound in neural-network verification
\citep{bunel2020branch}, and splitting works,
at a price that recurs at every level. The protocol here is the
cheap one: adaptive bisection along the widest dimension only, two
children per split, never a full grid. Under that favorable
protocol, on \textsc{ratd} just below the Chebyshev single-set
wall ($\ws$-scale $10.1$), Taylor needs 2 leaves ($3$ certified
runs, $2\times$ wall-clock) to cover what Chebyshev certifies as one
set; at scale $12.2$, 3 leaves versus 2; at $14.2$, 4 versus 3. Every
leaf is a full certified run, so matching Chebyshev costs Taylor
$1.4$ to $3\times$ the certified runs at these stresses, and the
full-grid worst case of $2^{n}$ leaves per level is what the cheap
protocol degenerates into when several directions stress at once. On the open-loop side the exchange rate is small where the
frontiers nearly coincide: on the spiral classifier of
Figure~\ref{fig:nnspiral} at $\epsilon=0.12$, single-set Taylor-4
leaves $24.7\%$ of the plane uncertified against $23.6\%$ for
Chebyshev-4, and best-axis $2$-splitting (cost $2\times$) takes
Taylor-4 to $22.1\%$, quartering (cost $4\times$) to $20.8\%$; the
basis is worth well under one doubling there, and the cost recurs at
every level, because Chebyshev can split too.

\begin{figure*}[t]
\centering
\includegraphics[width=\textwidth]{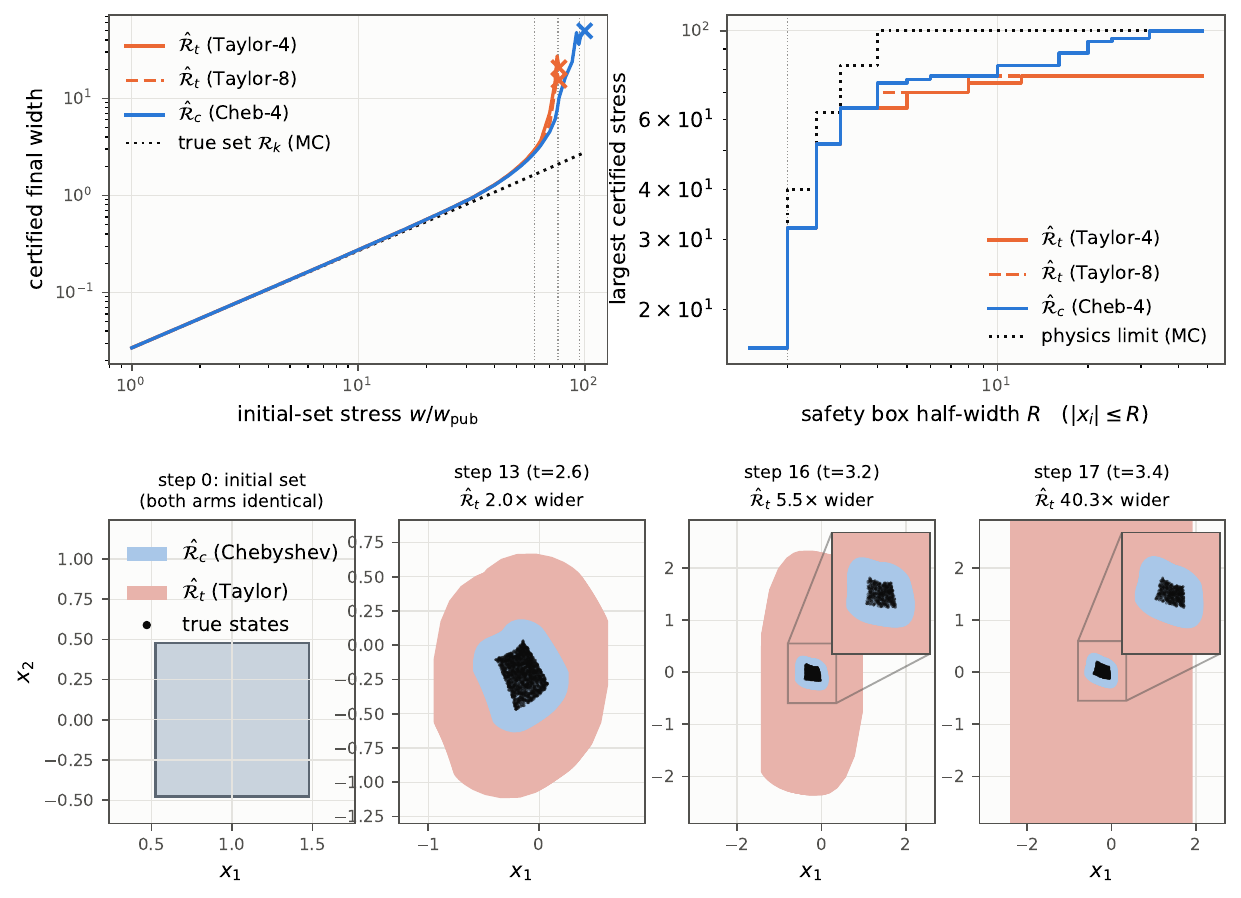}
\caption{The AINNCS TORA benchmark (sigmoid controller) under
initial-set stress $w$, in multiples of the published half-width.
\textbf{(a)}~Certified final widths of $\Rt$ and $\Rc$ against the
true set (dotted); crosses mark each method's last certified stress,
vertical dotted lines the bisected completion walls.
\textbf{(b)}~Safety frontier: largest certified stress for the box
$|x_i|\le R$; dotted line the physics limit, vertical dotted line
the published box $R=2$. \textbf{(c)}~The divergence mechanism on
the rational-damping loop of Section~\ref{sec:survival}, initial
half-width between the two walls (frame follows $\Rt$, insets
magnify $\Rc$, dots are true states): $\Rt$ balloons from
$2.0\times$ to $40\times\,\Rc$, fills the frame at $k=17$, and loses
its certificate at $k=18$; $\Rc$ hugs the truth for all 100 steps.
All configurations pass Monte-Carlo containment with zero
violations.}
\label{fig:ainncs}
\end{figure*}

The benchmarks so far are ours; the obvious question is what happens
on the community's.
We ran published instances of the ARCH-COMP AINNCS category
\citep{johnson2021arch} and the ReachNN suite
\citep{huang2019reachnn} under their published settings: TORA with the sigmoid
and $\tanh$ controllers (control period $0.5$, ten steps), ACC with
its $\tanh$ controller, and ReachNN B1 with its sigmoid controller.
Every method proves the published specification at the published
initial set, in the discrete-map semantics of
Section~\ref{sec:problem} and at the published control and
integration steps, with zero Monte-Carlo violations. At these initial sets $\Rt$
and $\Rc$ agree to eight significant digits: the community instances
leave the abstraction layer idle. The
discriminating question is how much more uncertainty each certificate
absorbs before it fails, and there the bases separate sharply
(Figure~\ref{fig:ainncs}).

\paragraph{Certified width under stress.}
Figure~\ref{fig:ainncs}(a) inflates the TORA initial box by a stress
factor $w$ and tracks the certified final width. All methods follow the
true set up to $w\approx46$. The Taylor-$4$ certificate then
inflates, to $1.6\times$ the Chebyshev width at $w=70$ and
$3.6\times$ at $w=75.5$, and from $w=82$ it returns \textsc{unknown};
doubling the Taylor degree no longer moves this wall (Taylor-$8$
fails at the same stress), while Chebyshev at degree $4$ still
certifies at $w=100$, the sweep maximum. The bisected completion
walls tell the same story ($76.2$ for Taylor at either degree against
$102.2$ for Chebyshev-$4$): with tight remainders on both sides the
Taylor wall is set by its truncation and the Chebyshev wall by the
carrier. On the $\tanh$ controller the last certified stresses are
$116$/$109$ (Taylor-$4$/-$8$) against $170$ (Chebyshev, sweep
maximum), and the completion walls $108$ against $181$.

\paragraph{Safety frontier.}
Figure~\ref{fig:ainncs}(b) asks the specification-level question: for
the safety box family $|x_i|\le R$, what is the largest certified
stress? Near the physics limit ($R\le2.5$, which includes the
published $|x_i|\le2$ box) all bases tie; the margin is consumed by
the true dynamics, and no abstraction can do better. From $R=4$ the
frontier forks, and the Chebyshev advantage grows to $1.30\times$
(sigmoid) and $1.56\times$ ($\tanh$) at $R=48$, in both cases with
Chebyshev censored at the sweep maximum, approaching the
completion-wall ratios of $1.34$ and $1.67$. The basis decides a safety verdict exactly
when the certificate, rather than the physics, is the binding
constraint.

\paragraph{Switching horizon.}
At stress $w=52$ with the sigmoid controller ($0.68$ of the
Taylor-$4$ completion wall) and an extended horizon, the certificates
survive $k=32$/$33$/$40$ control switches (Taylor-$4$/-$8$,
Chebyshev-$4$), and at $w=88$ with $\tanh$ ($0.81$ of its wall)
$k=17$/$18$/$25$: Chebyshev outlasts Taylor by $1.25$--$1.5\times$
at equal degree, and still outlasts Taylor at twice the degree. At
$w=40$ every method exceeds $640$ switches and the horizon does not
bind.

The mechanism is the same compounding everywhere. On TORA at
$w=82$, $\Rt^{(k)}$ and $\Rc^{(k)}$ differ by at most $1.1\times$
through $k=5$; the Taylor remainder then multiplies at each pass
through the switching surface, reaching $1.7\times$ at $k=7$,
$2.2\times$ at $k=8$, $3.6\times$ at $k=9$, and width $196$ at
$k=10$, past the divergence threshold, so the certificate is lost,
while $\Rc$ stays within width $16$ (a factor $8$ over the true
set $\mathcal{R}_k$) through all ten switches; Figure~\ref{fig:ainncs}(c) replays the
same collapse step by step on the rational-damping loop.

\paragraph{External anchor.} A final check concerns the strength
of the Taylor baseline: it is
at least as strong as what production tools ship, consistent with
the family-minimum remainder discipline of
Section~\ref{sec:method}, which production Taylor tools do not
use. On the pendulum initial-set race under matched
horizons, our Taylor method certifies $1.4$--$2.2\times$ \emph{larger}
initial sets than the Taylor-model tool TMJets (JuliaReach)
throughout the nonlinear regime $A\ge1$, noting that TMJets
certifies the continuous flow, a stronger object than our
discrete-map certificates (Section~\ref{sec:problem}); TMJets'
shrink-wrapping pays only in the near-linear regime.

\section{Scope and Limits}\label{sec:limits}

This section verifies the predicted null cases and reports the
misses.

\begin{table}[t]
\centering\footnotesize\setlength{\tabcolsep}{3.5pt}
\caption{Applicability map: predicted versus measured (real-axis
tie: bounded $w/\delta$).}
\label{tab:map}
\begin{tabular}{llll}
\toprule
scalar class & prediction & measured & verdict\\
\midrule
poles near axis & exponential sep. & walls $1.1$--$1.7\times$, flips & \textbf{win}\\
sharp gates ($\tau\!\downarrow$) & cap $\propto\tau$ & cap $+2\%$ at $N{=}32$ & \textbf{win}\\
entire ($\sin,\exp$) & $\le2^{N/(N+1)}$ & $1.07$--$1.16$ & constant\\
real-axis ($x^{-3/2}$) & bounded ratio & $1.02$ & tie\\
polynomial fields & bit-for-bit tie & max dev.\ $0.0$ (280 pairs) & tie\\
\bottomrule
\end{tabular}
\end{table}

Three boundaries complete the claim. \emph{(i) Ties are exact where
predicted} (Table~\ref{tab:map}); the remainder-constant asymmetry of
our own bounds (the Chebyshev families are up to an order of magnitude
more conservative than the Taylor Lagrange forms at low degree) means
reported separations are \emph{lower} bounds on the basis effect.
\emph{(ii) A carrier ceiling, not a physics ceiling.} At high degree
($N\ge8$) both bases on the gate family saturate against a common,
$\tau$-dependent wall. Monte-Carlo attribution shows the true reachable
sets remain bounded (width $\le1$) and ultimately contract to a point,
while both certified runs ratchet-explode once the carried box reaches
width $\approx1$: the ceiling belongs to the \emph{box representation}
(the wrapping ratchet of per-step re-boxing), not to the dynamics and
not to the basis. Removing it requires a matrix-valued hand-off
(Lohner-QR), an orthogonal upgrade and the subject of future work.
With tight remainders on both sides the ceiling already binds at
$N=4$: it compresses the closed-loop separation on the pendulum,
rational and gate families to $1.07$--$1.22$ (Table~\ref{tab:walls})
against a scalar factor of $2^{N}=16$, so the basis advantage is
largest exactly where the abstraction rather than the carrier is the
binding constraint: wide sets, sharp gates, and the scalar and
output-set level.
\emph{(iii) Failed predictions.} Pre-registered predictions that missed
are retained in the artifact: sharpening the gate did \emph{not} unlock
the high-$k$ separation (the carrier ceiling tracks
$\tau^{0.66\text{--}0.72}$ across both bases and degrees $4$ and $8$,
almost as fast as the pole height $\tau$), and the Taylor wall in the
rational family grows with $k$ rather than staying flat (death is a
growth race, not an instantaneous divergence event). A later
pre-registered round, run after the Taylor remainder families were
completed with the exact derivative members of
Section~\ref{sec:remainders}, missed in the other direction: the
closed-loop ratios fell from the $1.33$--$1.38$ of the first runs to
$1.09$--$1.22$ and the certified-area gaps on the spiral classifiers
from ten to twenty points to five to seven, because the earlier
Taylor baseline had carried remainders $9$ to $30\times$ above the
true error on the gates. All numbers reported here are from the
completed families. Each miss sharpened the mechanism picture above.

One pilot artifact
is retracted with its mechanism named: with the exact-range
fallback active, a diverged Taylor layer is silently replaced by the
true per-neuron range, which censors the basis comparison the same
way the box carrier censors walls; the open-loop ablation numbers of
Section~\ref{sec:nnwindow} are therefore measured raw, and the
fallback is reported as a separate design choice.


\section{Conclusion}\label{sec:conclusion}

Reachability's universal Taylor default applies a pointwise
expansion to a set-sized argument. On most classical dynamics the
mismatch never surfaces, and our provable-tie results say exactly
when it cannot; on the sharply gated and rational dynamics modern
control is producing, it caps the certificate precisely on the
switching surface.
The fix is one changed operation, Chebyshev coefficients with
closed-form sound remainders inside the unchanged polynomial-zonotope
pipeline, and its effect is governed by quantitative laws: for
Taylor a certified wall at the singularity distance, for Chebyshev
geometric convergence at linearly growing degree, and the worst
case located directly above the switch. With a sound
softmax transformer the same machinery reaches the gates of learned
controllers. The map of what a basis change cannot improve (polynomial
ties, entire-function constants, the box-carrier ceiling of
Section~\ref{sec:limits}) is itself a
result: it locates the next bottleneck.

\bibliographystyle{IEEEtranN}
\bibliography{refs}

\section{Reproducibility}\label{app:repro}
All computations use IEEE-754 double precision, and every certified
radius includes an explicit rounding-error inflation. A
per-operation rounding analysis in the standard $(1+\delta)$ model
has not been carried out; machine-precision soundness therefore
rests on the inflated radii, and formalizing it is future work. The
complete implementation, the drivers for every experiment, and the
pre-registered prediction notes (including the misses of
Section~\ref{sec:limits}) accompany the paper.

\end{document}